\documentclass{elsarticle}
\usepackage{{booktabs}}
\usepackage{subcaption}
\usepackage{caption}
\usepackage{multirow}
\usepackage{graphicx}
\usepackage{hyperref}
\usepackage{color}
\usepackage{xcolor}
\definecolor{grey}{RGB}{77,77,77}
\definecolor{myred}{RGB}{234,107,102}

\usepackage{tikz}

\begin{document}

\begin{frontmatter}

\title{LLM4Jobs: Unsupervised occupation extraction and standardization leveraging Large Language Models
}

\author{Nan Li}
\ead{nan.li@ugent.be}
\author{Bo Kang}
\ead{bo.kang@ugent.be}
\author{Tijl De Bie}
\ead{tijl.debie@ugent.be}

\affiliation{organization={IDLAB - Department of Electronics and Information Systems (ELIS), Ghent University},
postcode={9000},
city={Ghent},
country={Belgium}}

\begin{abstract}
Automated occupation extraction and standardization from free-text job postings and resumes are crucial for applications like job recommendation and labor market policy formation. This paper introduces LLM4Jobs, a novel unsupervised methodology that taps into the capabilities of large language models (LLMs) for occupation coding. LLM4Jobs uniquely harnesses both the natural language understanding and generation capacities of LLMs. Evaluated on rigorous experimentation on synthetic and real-world datasets, we demonstrate that LLM4Jobs consistently surpasses unsupervised state-of-the-art benchmarks, demonstrating its versatility across diverse datasets and granularities. As a side result of our work, we present both synthetic and real-world datasets, which may be instrumental for subsequent research in this domain. Overall, this investigation highlights the promise of contemporary LLMs for the intricate task of occupation extraction and standardization, laying the foundation for a robust and adaptable framework relevant to both research and industrial contexts.
\end{abstract}

\end{frontmatter}
\section{Introduction}\label{sec:intro}
Job descriptions and resumes encapsulate invaluable information, notably concerning occupations and associated skills. An ``occupation'' refers to an individual's professional role, encompassing the duties and responsibilities specific to that position. The challenge lies in accurately classifying these occupations, given the intricate nature of job roles and their descriptions.

Automatic occupation extraction aims to extract occupation information from textual data, eliminating the need for manual intervention. Occupation standardization or coding then maps the free-styled occupation terms onto a certain standard taxonomy, such as the ``International Standard Classification of Occupations (ISCO)''\cite{isco} and the ``European Skills, Competences, Qualifications and Occupations'' (ESCO) standard\cite{esco}. Such coding facilitates various types of downstream usage, such as job recommendation, occupational health studies\cite{wan2023automated}, labour market policies design and evaluation\cite{boselli2017using}. %

Current state-of-the-art methods for occupation extraction and standardization (OES) predominantly employ \emph{supervised} techniques, framing the task as a text classification problem. A majority of these studies rely solely on job titles as features \cite{bethmann2014automatic,boselli2017using,boselli2018wolmis,boselli2018classifying}, despite the titles' inherent inconsistencies and noise\cite{ikudo2019occupational}. A few researchers incorporate both job titles and descriptions\cite{varelas2022employing}, necessitating cumbersome natural language processing (NLP) techniques for effective data preprocessing. This complexity arises from the diverse and embellished language often found in job postings, which, while intended to attract potential candidates, diverges from standardized occupation descriptions and adds details about perks and prerequisites. 
Specific duty descriptions reported by people, which align more closely with actual occupations, have been explored\cite{russ2016computer}, but such data sources are severely limited in their accessibility and use.

On the other hand, \emph{unsupervised} learning approaches remain in their infancy. Early attempts include rule-based methods \cite{cascot} and key-word search \cite{laborr}. While these methods eliminate the need for labor-intensive human annotation, they have unsatisfying accuracy \cite{wan2023automated} and are constrained by language and taxonomy dependencies. 

With the advent of GPT-style decoder-only \emph{large language models} (LLMs), there's a newfound potential to address these challenges. 
These models, typified by platforms like ChatGPT, amplify conversational interfaces' benefits, facilitating non-programmer domain experts to engage seamlessly. 
However, current LLMs have some limitations that make them hard to apply directly for OES. 
Directly prompting a job posting for occupation codes might well be answered with non-existent codes due to the \emph{hallucination} problem \cite{dziri2022origin,ji2023survey}. 
Thus, to improve the accuracy, occupation code definitions should be fed into the LLM as context. 
Yet, handling \emph{lengthy prompts} of hundreds of pages is not straightforward. 
Even if such documents were somehow prompted (e.g. by chunk), model predictions would be \emph{inefficient}, making it infeasible for online usage. 
Furthermore, using powerful but \emph{costly and close-sourced} models such as GPT-4 from OpenAI would hinder exploration and evaluation of possible solutions.

To address the aforementioned issues, we introduce \textbf{LLM4Jobs}, an \emph{unsupervised two-phase} method uniquely combines the natural language understanding and generation abilities of advanced LLMs to ensure accuracy, efficiency, and cost-effectiveness. 

In the first phase, \textbf{LLM4Jobs} computes embeddings for the occupation taxonomy elements. Each occupation's textual description is processed through the LLM to obtain token-level embeddings, then the mean of these token embeddings is taken as the final embedding for the occupation, serving as a foundational reference for subsequent matching operations. 
In the second query phase, for any incoming free-text job description or user profile, an optional summarization step can be employed to distill the essence of the text, then the same embedding technique is applied to get the vectorized representation of the query. This representation is subsequently matched against the precomputed embeddings of the occupation taxonomy using a \emph{rapid} vector similarity search, allowing for the precise retrieval of the most fitting occupation codes.

Furthermore, our reliance on the most recent \emph{open-source} LLMs ensures minimal deployment costs, especially in academic settings.
A notable advantage of LLM4Jobs lies in its \emph{adaptability}. It is agnostic to specific taxonomies or languages, offering flexibility concerning the backbone model and adaptability for other applications.

\paragraph{Contributions}\label{sec:contributions}
\begin{enumerate}
\item We introduce \emph{LLM4Jobs}, the first framework that leverages decoder-only large language model for occupation extraction and standardization tasks. This approach uniquely harnesses the generative capabilities of LLMs to aid text representation.
\item We investigated and confirmed the utility of LLM-based summarization of job descriptions and resumes, with the purpose of making occupation information more salient in lengthy texts, thus improving the accuracy of LLM4Jobs.
\item We introduce two novel synthetic occupation coding datasets generated using GPT-4, and a dataset crafted from manually annotated real-world job postings. These \emph{datasets} aim to serve as benchmarks for future research in the domain (\url{https://github.com/aida-ugent/Occupation_coding_datasets}).
\item Comprehensive \emph{experiments} showcase the effectiveness of LLM4Jobs against contemporary unsupervised baselines. We evaluate the utility of various open-source LLaMA-based LLMs and different LLM4Jobs configurations, offering insights into its versatility and performance.
\item Our open-source \emph{implementation} of LLM4Jobs is flexible and well-suited for adaptation to diverse applications. The code is publicly accessible at \url{https://github.com/aida-ugent/SkillGPT}.
\end{enumerate}

\section{Methodology}\label{sec:method}
We will first briefly define the problem as it is commonly conceived in Sec.~\ref{sec:prob_def}. Then, in Sec.~\ref{sec:LLM4Jobs_framework}, we will explain the approach we adopted for solving it in LLM4Jobs.
\subsection{Problem Definition}\label{sec:prob_def}
The essence of occupation coding lies in accurately mapping a job description to its corresponding standardized occupation code(s). This task can be elegantly captured as a selection and ranking problem. Given a document \( d \) (representing the job description) and a set \( C \) of standardized occupation codes, the first step is to associate \( d \) with a subset of codes \( C' \) that may plausible describe the job role portrayed in \( d \). Mathematically, this first selection step can be described using the function: $f: d \rightarrow C'$.

Then a ranked list \( L \) of occupation codes is produced by sorting \( C' \) such that, for any two codes \( c_i \) and \( c_j \) in \( L \), \( c_i \) is ranked higher than \( c_j \) if \( c_i \) is more pertinent to \( d \) than \( c_j \). 

\subsection{LLM4Jobs Framework}\label{sec:LLM4Jobs_framework}

\begin{figure}[!h]
\includegraphics[width=\textwidth]{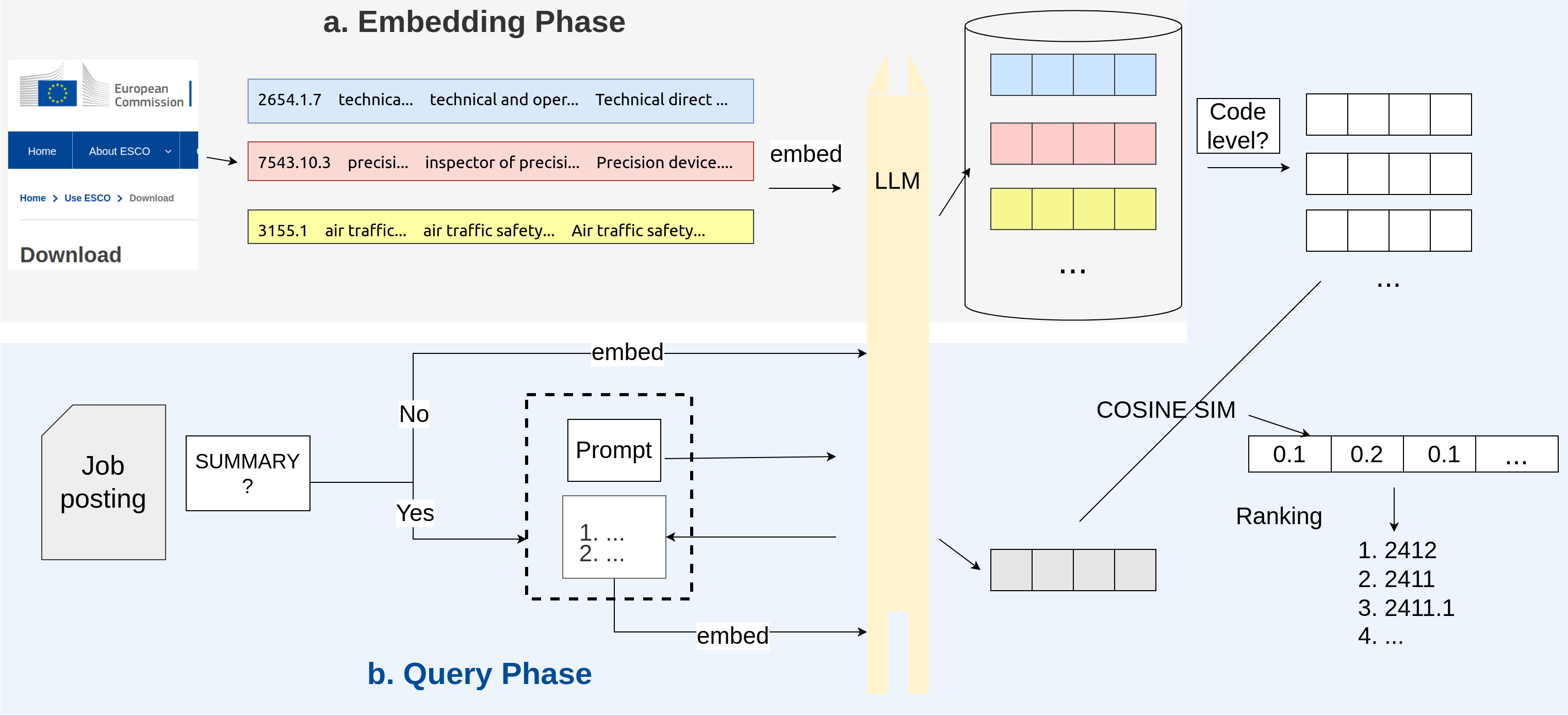}
\caption{LLM4Jobs framework.}
\label{fig:method}
\end{figure}

The overall approach for solving this problem adopted in LLM4Jobs is illustrated in Fig.~\ref{fig:method}. LLM4Jobs has two primary phases: the \textit{Embedding Phase} (as depicted in Fig.~\ref{fig:method}a) and the \textit{Query Phase} (illustrated in Fig.~\ref{fig:method}b). During the Embedding phase, the occupation descriptions from the taxonomy are embedded using an LLM. During the Query phase occupations are ranked based on the similarity of their embedding to an embedding of the query document (a job description or resume) with the same LLM. Next we describe each of these steps in some greater detail.

\subsubsection{Embedding Phase}
This phase is central to generating embeddings for the occupation taxonomy database. By ensuring these embeddings are both precise and efficiently retrievable, the foundation is laid for the subsequent query operations. 

To calculate the embeddings, each occupation's textual labels and definitions are first concatenated. This combined text is then input into a state-of-the-art LLM. The LLM produces embeddings for each token in the sentence. The mean of these token embeddings is taken to derive a single vectorized representation for the entire occupation description\footnote{We also experimented with using the embedding of the last token as a representation of the sentence, but this approach yielded inferior results compared to the mean embedding.}. This representation is subsequently stored for future reference.

\subsubsection{Query Phase}
The query phase is a multi-step process designed to transform a given document, typically a job posting, into a ranked list of relevant occupation codes. This phase is further broken down as follows:

\begin{enumerate}
    \item \textbf{Query Preparation}:
    \begin{itemize}
        \item Document summarization: This is an optional step in which the query text is summarized (details in \ref{app:generation_hyperparam}). The purpose is to distill the essence of the job description, which we hypothesized would result in more accurate occupation classifications. As we validate in the experiments, it is particularly useful for long real-world job descriptions that may contain substantial portions of text that are largely uninformative of the occupation.
        \item Code Granularity Selection: This step determines the desired granularity of the occupation code, adaptable to different taxonomies. For instance, users can specify the level four of the ESCO/ISCO code.
    \end{itemize}
    
    \item \textbf{Candidate Retrieval}:
    \begin{itemize}
        \item Document Embedding: The system transforms the query document into an embedding representation using the same method as for embedding the taxonomy, suitable for computing similarities.
        \item Code Embedding Retrieval: Depending on the previously chosen granularity, relevant occupation code embeddings are retrieved. The choice of embedding is guided by the selected \emph{mapping strategy} (see more below).
        \item Similarity Computation: The cosine similarity between the query and code embeddings is calculated, establishing the relevance of each code to the document.
    \end{itemize}
    
    \item \textbf{Ranking}: Based on computed similarity scores, the system ranks the codes, delivering the top-k most relevant occupation codes for the given document.
\end{enumerate}

A further justification for the summarization option is as follows. Online job postings often employ a rich and promotional lexicon to appeal to potential candidates. While this style can be effective for recruitment, it tends to obfuscate the core responsibilities and qualifications associated with the job, complicating the task of accurate occupation classification. To address this, we leverage the generative capabilities of LLMs. By instructing the backbone LLM to extract and summarize the essential duties and responsibilities from the job posting, we can obtain a condensed representation that aligns more closely with standardized occupation descriptions. This streamlined text is then better positioned for similarity assessments against the embedded standard descriptions of occupations.

The hierarchical nature of occupation codes necessitates specific strategies to handle different granularity levels. Viable strategies include:
\begin{enumerate}
        \item \textbf{Truncation}: Retrieve the most granular codes and truncate them to the desired level. For example, the ESCO occupation code 4222.1.1 (live chat operator) could be truncated to the level four code 4222 (contact center information clerks) or level 3 code 422 (client information workers).
        \item \textbf{Direct Mapping}: Directly use the descriptions of the codes at the target level of granularity codes.
        \item \textbf{Clustering}: For each code at the target level of granularity, aggregate the embeddings of its most granular subordinate codes, and use their mean as the representative embedding.
\end{enumerate}

Subsequent sections (specifically, Sec.~\ref{sec:rq2_results}) elucidate the impact of the summarization approach and the mapping strategy choice across diverse datasets.

\section{Experiments}\label{sec:experiments}
To validate the efficacy of LLM4Jobs, we devised a series of experiments addressing the following research questions:
\begin{enumerate}
\item How does LLM4Jobs perform relative to existing unsupervised methods in the domain of occupation extraction and standardization? \label{rq1}
\item What combinations of summarization and mapping strategies yield effective results across different settings in LLM4Jobs? \label{rq2} %
\item What is the performance differential when employing smaller LLMs as the backbone? \label{rq3}
\end{enumerate}
The subsequent sections detail the experimental setup, starting with the foundational settings in Sec.~\ref{sec:settings}. We then outline the datasets utilized, both synthetic and manually-annotated, in Sec.~\ref{sec:datasets}. Finally, Sec.~\ref{sec:results} presents the results from these experiments, offering insights into the advantages and limitations of LLM4Jobs and its various configurations.

\subsection{Settings}\label{sec:settings}%
\subsubsection{RQ\ref{rq1}: Baselines}\label{sec:baselines}
For an empirical assessment, we compare LLM4Jobs with two renowned unsupervised methodologies:
\begin{itemize}
\item \textbf{CASCOT}\cite{cascot}: a software recognized for its proficiency in ISCO coding (ESCO occupation codes at level-4). A recent study\cite{wan2023automated} commends CASCOT as the leading tool available to the public for automatic ISCO coding.
\item \textbf{GPT-4 Zero-shot}: Harnessing GPT-4's inherent capabilities, we prompted it, via OpenAI's API, asking it for the ten most plausible ESCO/ISCO occupation codes for a given job description. The specifics of these prompts are elaborated in \ref{app:gpt4_prompts}.
\end{itemize}

For the benchmark against baseline methods, we configured LLM4Jobs in a specific manner. The Vicuna-33b was chosen as the backbone LLM, driven by our hypothesis favoring larger LLMs and constrained by available resources. For predictions at less granular code levels, we employed truncation mapping, as it aligns with the methodology of the baselines. An adaptive summarization approach was incorporated, triggering summarization only for documents exceeding an empirically determined threshold of 300 words (details in \ref{app:generation_hyperparam}). This choice was motivated by efficiency considerations and the observation that longer job postings tend to be more verbose, benefiting from distillation.

\subsubsection{RQ\ref{rq2}: Variations within LLM4Jobs}\label{sec:variants}
To discern the impact of summarization, three methodologies were assessed: no summarization, universal document summarization, and adaptive summarization based on document length.

The influence of the mapping strategy was evaluated by comparing three techniques for predicting less granular ESCO/ISCO occupation codes. While the maximum granularity of ESCO codes varies (spanning from level 5 to 8, termed as level-5+), they all extend the 4-level ISCO codes. For predictions at levels 3 and 4, LLM4Jobs offers the following alternatives:
\begin{enumerate}
\item \textbf{Truncation}: Utilizing all level-5+ codes as potential matches and truncating predictions to the desired level.
\item \textbf{Direct Mapping}: Retrieving embeddings solely of level-3/4 codes.
\item \textbf{Clustering}: For each level-3/4 code, embeddings of all subordinate leaf codes (level-5+ codes sharing the same 3/4-digit prefix) are aggregated, with their mean representing the parent code.
\end{enumerate}
The three summarization approaches combined with the three levels considered results in nine potential LLM4Jobs configurations.

\subsubsection{RQ\ref{rq3}: Opting for the Backbone Model}\label{sec:backboneLLMs}
While we anticipate superior performance from larger models, we sought to validate this presumption for our specific task and quantify the performance degradation with smaller models.

Several open-sourced large language models were scrutinized: Vicuna-7b, Vicuna-13b, Vicuna-33b, LLaMA-2-13b, and LLaMA-2-13b-chat. The Vicuna series, evolved by refining LLaMA\cite{touvron2023llama}, has been hailed for outshining its LLaMA predecessors \cite{zheng2023judging}. These models, to our knowledge, rank among the foremost open-source GPT-like models. The numerical tags (7, 13, 33) represent their parameter count (in billions). Our choice of model size was constrained by computational limitations. Additionally, the LLaMA-2 series was chosen for its latest iteration and unrestricted commercial applicability.

\subsubsection{Evaluation Metrics}\label{sec:metrics}
Performance was gauged using three established metrics:
\begin{enumerate}
\item HR@k: Hit ratio at k, the proportion of instances where the accurate item is featured in the top-k predictions.
\item MRR@k: Mean reciprocal rank at k, which quantifies the rank of the first relevant item found in the predicted list, also known as average reciprocal hit ratio (ARHR).
\item NDCG@k: Normalized discounted cumulative gain at k, which measures the quality of the ranked list, weighing the positions of relevant items.
\end{enumerate}

For our specific problem, there is only one occupation corresponding to each data point, i.e. only one relevant item, and we assign relevant score 1 to the true label and 0 to others for simplicity. Therefore, HR@1, MRR@1 and NDCG@1 are equivalent and simply equal to the accuracy.

\subsection{Datasets}\label{sec:datasets}
In the absence of public benchmark datasets, we forged two synthetic datasets with pre-assigned labels and a manually annotated real-world dataset.

\subsubsection{Synthetic datasets}
We randomly sampled 1000 ESCO occupation codes. For each code, we used OpenAI's API to prompt the latest GPT-4 model to generate a corresponding job posting. To enhance the challenge and realism of the dataset, 500 of these postings were generated with explicit instructions to deviate the job title from the standard occupation label. These postings form the ``GenHard'' dataset, while their counterparts constitute the ``GenEasy'' dataset. Further details are available in \ref{app:gpt4_prompts}.

\subsubsection{Real world dataset}
We sampled and annotated 100 real public job postings from the American job portal, Indeed\footnote{\url{https://www.indeed.com/}},detailed in \ref{app:real_data}. Each job posting has two annotators and the disagreements are resolved by a third annotator. Using the final consensus codes as the ground truth, the average accuracy of human annotators against the final consensus annotations is 0.83. Note that this may suggest as a hard upper bound on the accuracy we can expect any automated system to achieve.

\section{Results}\label{sec:results}

\begin{table}[!h]
\begin{footnotesize}
\setlength{\tabcolsep}{1pt}
\caption{Comparison of performance of all methods on the \textbf{real-world} dataset. The backbone model of LLM4Jobs is Vicuna-33b. The LLM4Jobs variant followed by $\dagger$ is the default strategy used for comparing with the baselines, where the highest scores are underlined (addressing RQ\ref{rq1}). Between the variants of LLM4Jobs, the highest scores are boldfaced (addressing RQ\ref{rq2}). CASCOT predicts only level-4 ESCO codes, i.e. ISCO codes, hence no results for level-5+.}
\label{tab:real}
\begin{tabular}{llllrrrrrrr}
\toprule
 &  &  &  &HR@1 &  \multicolumn{1}{c}{@5} & \multicolumn{1}{c}{@10} & MRR@5 &\multicolumn{1}{c}{@10} & NDCG@5 & \multicolumn{1}{c}{@10} \\
Level & Method & Summary & Mapping &  &  &  &  &  &  &  \\
\midrule
\multirow[c]{11}{*}{3} & \multirow[c]{9}{*}{LLM4Jobs} & \multirow[c]{3}{*}{adaptive} & cluster & 0.300 & 0.690 & 0.780 & 0.456 & 0.469 & 0.515 & 0.545 \\
 &  &  & direct & 0.360 & 0.720 & 0.850 & 0.497 & 0.514 & 0.553 & 0.595 \\
 &  &  & truncation $\dagger$ & 0.390 & \underline{\textbf{0.800}} & \underline{0.860} & \underline{\textbf{0.538}} & \underline{0.546} & \underline{\textbf{0.603}} & \underline{\textbf{0.622}} \\
 &  & \multirow[c]{3}{*}{all} & cluster & 0.320 & 0.690 & 0.780 & 0.469 & 0.482 & 0.524 & 0.555 \\
 &  &  & direct & \textbf{0.430} & 0.720 & 0.850 & 0.534 & \textbf{0.552} & 0.580 & \textbf{0.622} \\
 &  &  & truncation & 0.380 & 0.760 & \textbf{0.870} & 0.520 & 0.535 & 0.580 & 0.615 \\
 &  & \multirow[c]{3}{*}{no} & cluster & 0.270 & 0.670 & 0.760 & 0.412 & 0.424 & 0.476 & 0.505 \\
 &  &  & direct & 0.280 & 0.680 & 0.810 & 0.427 & 0.443 & 0.489 & 0.530 \\
 &  &  & truncation & 0.270 & 0.720 & 0.810 & 0.440 & 0.453 & 0.510 & 0.540 \\
 & CASCOT & - & truncation & 0.130 & 0.400 & 0.590 & 0.226 & 0.250 & 0.269 & 0.329 \\
 & GPT4 & - & truncation & \underline{0.510} & 0.550 & 0.570 & 0.527 & 0.529 & 0.533 & 0.539 \\
\midrule
\multirow[c]{11}{*}{4} & \multirow[c]{9}{*}{LLM4Jobs} & \multirow[c]{3}{*}{adaptive} & cluster & 0.320 & 0.580 & 0.650 & 0.435 & 0.444 & 0.472 & 0.494 \\
 &  &  & direct & 0.390 & \textbf{0.710} & 0.770 & 0.508 & 0.516 & 0.558 & 0.578 \\
 &  &  & truncation $\dagger$ & \underline{0.320} & \underline{0.570} & \underline{0.650} & \underline{0.409} & \underline{0.420} & \underline{0.449} & \underline{0.475} \\
 &  & \multirow[c]{3}{*}{all} & cluster & 0.310 & 0.590 & 0.660 & 0.430 & 0.439 & 0.470 & 0.493 \\
 &  &  & direct & \textbf{0.430} & 0.700 & \textbf{0.810} & \textbf{0.530} & \textbf{0.545} & \textbf{0.573} & \textbf{0.608} \\
 &  &  & truncation & 0.300 & 0.530 & 0.660 & 0.389 & 0.407 & 0.425 & 0.467 \\
 &  & \multirow[c]{3}{*}{no} & cluster & 0.240 & 0.560 & 0.660 & 0.365 & 0.378 & 0.414 & 0.446 \\
 &  &  & direct & 0.230 & 0.600 & 0.670 & 0.358 & 0.367 & 0.418 & 0.440 \\
 &  &  & truncation & 0.230 & 0.520 & 0.630 & 0.338 & 0.353 & 0.383 & 0.419 \\
 & CASCOT & - & - & 0.100 & 0.310 & 0.500 & 0.174 & 0.199 & 0.208 & 0.269 \\
 & GPT4 & - & truncation & 0.280 & 0.390 & 0.420 & 0.319 & 0.322 & 0.337 & 0.346 \\
 \midrule
\multirow[c]{5}{*}{5+} & \multirow[c]{3}{*}{LLM4Jobs} & adaptive $\dagger$ & - & \underline{\textbf{0.220}} & \underline{\textbf{0.480}} & \underline{0.550} & \underline{\textbf{0.319}} & \underline{\textbf{0.328}} & \underline{\textbf{0.359}} & \underline{\textbf{0.382}} \\
 &  & all & - & 0.190 & 0.450 & \textbf{0.560} & 0.295 & 0.309 & 0.334 & 0.369 \\
 &  & no & - & 0.170 & 0.400 & 0.500 & 0.254 & 0.267 & 0.290 & 0.322 \\
 & CASCOT & - & - & - & - & - & - & - & - & - \\
 & GPT4 & - & - & 0.070 & 0.100 & 0.140 & 0.079 & 0.083 & 0.084 & 0.096 \\
\bottomrule
\end{tabular}
\end{footnotesize}
\end{table}
\begin{table}[!h]
\begin{footnotesize}
\centering
\setlength{\tabcolsep}{1pt}
\caption{Comparison of performance of all methods on the \textbf{GenEasy} dataset. The backbone model of LLM4Jobs is Vicuna-33b. The LLM4Jobs variant followed by $\dagger$ is the default strategy used for comparing with the baselines, where the highest scores are underlined (addressing RQ\ref{rq1}). Between the variants of LLM4Jobs, the highest scores are boldfaced (addressing RQ\ref{rq2}). CASCOT predicts only level-4 ESCO codes, i.e. ISCO codes, hence no results for level-5+. Since the document lengths are all below 300 words, the adaptive summarization is the same with no summary.}
\label{tab:easy}
\begin{tabular}{lcclrrrrrrr}
\toprule
 &  &  &  &HR@1 &  \multicolumn{1}{c}{@5} & \multicolumn{1}{c}{@10} & MRR@5 &\multicolumn{1}{c}{@10} & NDCG@5 & \multicolumn{1}{c}{@10} \\
Level & Method & Summary & Mapping &  &  &  &  &  &  &  \\
\midrule
\multirow[c]{8}{*}{3} & \multirow[c]{6}{*}{LLM4Jobs} & \multirow[c]{3}{*}{adaptive} & cluster & 0.582 & 0.860 & 0.926 & 0.695 & 0.704 & 0.737 & 0.758 \\
 &  &  & direct & 0.306 & 0.670 & 0.798 & 0.433 & 0.451 & 0.491 & 0.534 \\
 &  &  & truncation $\dagger$ & \underline{\textbf{0.724}} & \underline{\textbf{0.920}} & \underline{\textbf{0.964}} & \underline{\textbf{0.806}} & \underline{\textbf{0.812}} & \underline{\textbf{0.835}} & \underline{\textbf{0.849}} \\
 &  & \multirow[c]{3}{*}{all} & cluster & 0.564 & 0.838 & 0.900 & 0.676 & 0.685 & 0.717 & 0.737 \\
 &  &  & direct & 0.286 & 0.668 & 0.804 & 0.425 & 0.444 & 0.485 & 0.530 \\
 &  &  & truncation & 0.708 & 0.902 & 0.954 & 0.785 & 0.792 & 0.814 & 0.831 \\
 & CASCOT & - & truncation & 0.380 & 0.750 & 0.846 & 0.519 & 0.533 & 0.577 & 0.609 \\
 & GPT4 & - & truncation & 0.476 & 0.540 & 0.550 & 0.502 & 0.503 & 0.512 & 0.515 \\
\midrule
\multirow[c]{8}{*}{4} & \multirow[c]{6}{*}{LLM4Jobs} & \multirow[c]{3}{*}{adaptive} & cluster & 0.572 & 0.820 & 0.894 & 0.676 & 0.686 & 0.712 & 0.737 \\
 &  &  & direct & 0.366 & 0.670 & 0.756 & 0.480 & 0.492 & 0.527 & 0.556 \\
 &  &  & truncation $\dagger$ & \underline{\textbf{0.692}} & \underline{\textbf{0.902}} & \underline{\textbf{0.946}} & \underline{\textbf{0.777}} & \underline{\textbf{0.783}} & \underline{\textbf{0.809}} & \underline{\textbf{0.823}} \\
 &  & \multirow[c]{3}{*}{all} & cluster & 0.554 & 0.802 & 0.868 & 0.651 & 0.660 & 0.689 & 0.710 \\
 &  &  & direct & 0.380 & 0.670 & 0.768 & 0.487 & 0.501 & 0.533 & 0.565 \\
 &  &  & truncation & 0.676 & 0.868 & 0.924 & 0.750 & 0.758 & 0.780 & 0.798 \\
 & CASCOT & - & - & 0.298 & 0.660 & 0.762 & 0.432 & 0.446 & 0.489 & 0.522 \\
 & GPT4 & - & truncation & 0.220 & 0.268 & 0.286 & 0.237 & 0.240 & 0.245 & 0.251 \\
\midrule
\multirow[c]{4}{*}{5+} & \multirow[c]{2}{*}{LLM4Jobs} & adaptive $\dagger$ & - & \underline{\textbf{0.590}} & \underline{\textbf{0.818}} & \underline{\textbf{0.880}} & \underline{\textbf{0.683}} & \underline{\textbf{0.691}} & \underline{\textbf{0.717}} & \underline{\textbf{0.737}} \\
 &  & all & - & 0.558 & 0.780 & 0.846 & 0.647 & 0.656 & 0.681 & 0.702 \\
 & CASCOT & - & - & - & - & - & - & - & - & - \\
 & GPT4 & - & - & 0.014 & 0.038 & 0.054 & 0.023 & 0.025 & 0.027 & 0.032 \\
\bottomrule
\end{tabular}
\end{footnotesize}
\end{table}

\begin{table}[!h]
\begin{footnotesize}
\centering
\setlength{\tabcolsep}{1pt}
\caption{Comparison of performance of all methods on the \textbf{GenHard} dataset. The backbone model of LLM4Jobs is Vicuna-33b. The LLM4Jobs variant followed by $\dagger$ is the default strategy used for comparing with the baselines, where the highest scores are underlined (addressing RQ\ref{rq1}). Between the variants of LLM4Jobs, the highest scores are boldfaced (addressing RQ\ref{rq2}). CASCOT predicts only level-4 ESCO codes, i.e. ISCO codes, hence no results for level-5+. Since the document lengths are all below 300 words, the adaptive summarization is the same with no summary.}
\label{tab:hard}
\begin{tabular}{lcclrrrrrrr}
\toprule
 &  &  &  &HR@1 &  \multicolumn{1}{c}{@5} & \multicolumn{1}{c}{@10} & MRR@5 &\multicolumn{1}{c}{@10} & NDCG@5 & \multicolumn{1}{c}{@10} \\
Level & Method & Summary & Mapping &  &  &  &  &  &  &  \\
\midrule
\multirow[c]{8}{*}{3} & \multirow[c]{6}{*}{LLM4Jobs} & \multirow[c]{3}{*}{adaptive} & cluster & 0.500 & 0.834 & 0.892 & 0.637 & 0.645 & 0.687 & 0.706 \\
 &  &  & direct & 0.210 & 0.574 & 0.738 & 0.337 & 0.359 & 0.395 & 0.449 \\
 &  &  & truncation $\dagger$ & \underline{0.576} & \underline{\textbf{0.850}} & \underline{\textbf{0.916}} & \underline{0.683} & \underline{0.692} & \underline{0.725} & \underline{0.747} \\
 &  & \multirow[c]{3}{*}{all} & cluster & 0.470 & 0.814 & 0.890 & 0.611 & 0.621 & 0.662 & 0.687 \\
 &  &  & direct & 0.206 & 0.566 & 0.762 & 0.332 & 0.359 & 0.390 & 0.454 \\
 &  &  & truncation & \textbf{0.584} & \textbf{0.850} & 0.904 & \textbf{0.690} & \textbf{0.697} & \textbf{0.730} & \textbf{0.748} \\
 & CASCOT & - & truncation & 0.224 & 0.532 & 0.709 & 0.328 & 0.352 & 0.378 & 0.435 \\
 & GPT4 & - & truncation & 0.354 & 0.434 & 0.452 & 0.385 & 0.388 & 0.397 & 0.403 \\
\midrule
\multirow[c]{8}{*}{4} & \multirow[c]{6}{*}{LLM4Jobs} & \multirow[c]{3}{*}{adaptive} & cluster & 0.490 & 0.782 & 0.876 & 0.607 & 0.620 & 0.651 & 0.682 \\
 &  &  & direct & 0.270 & 0.572 & 0.716 & 0.383 & 0.402 & 0.430 & 0.477 \\
 &  &  & truncation $\dagger$ & \underline{\textbf{0.532}} & \underline{\textbf{0.806}} & \underline{\textbf{0.886}} & \underline{\textbf{0.638}} & \underline{\textbf{0.648}} & \underline{\textbf{0.680}} & \underline{\textbf{0.706}} \\
 &  & \multirow[c]{3}{*}{all} & cluster & 0.438 & 0.780 & 0.856 & 0.569 & 0.579 & 0.621 & 0.646 \\
 &  &  & direct & 0.290 & 0.624 & 0.712 & 0.412 & 0.423 & 0.465 & 0.493 \\
 &  &  & truncation & 0.528 & 0.802 & 0.854 & 0.633 & 0.640 & 0.675 & 0.692 \\
 & CASCOT & - & - & 0.149 & 0.426 & 0.595 & 0.242 & 0.265 & 0.287 & 0.342 \\
 & GPT4 & - & truncation & 0.180 & 0.242 & 0.262 & 0.203 & 0.205 & 0.212 & 0.219 \\
\midrule
\multirow[c]{4}{*}{5+} & \multirow[c]{2}{*}{LLM4Jobs} & adaptive $\dagger$ & - & \underline{\textbf{0.412}} & \underline{\textbf{0.680}} & \underline{\textbf{0.786}} & \underline{\textbf{0.512}} & \underline{\textbf{0.526}} & \underline{\textbf{0.554}} & \underline{\textbf{0.588}} \\
 &  & all & - & 0.378 & 0.658 & 0.750 & 0.479 & 0.492 & 0.524 & 0.554 \\
 & CASCOT & - & - & - & - & - & - & - & - & - \\
 & GPT4 & - & - & 0.002 & 0.016 & 0.032 & 0.006 & 0.008 & 0.008 & 0.014 \\
\bottomrule
\end{tabular}
\end{footnotesize}
\end{table}

Tables \ref{tab:real}, \ref{tab:easy}, and \ref{tab:hard} present the performance comparison of LLM4Jobs with baseline methods across three datasets: real-world, GenEasy, and GenHard. The default configuration of LLM4Jobs, denoted with a $\dagger$, serves to address RQ\ref{rq1}.

\begin{figure}
\raggedleft
\begin{subfigure}{.49\textwidth}
    \centering
    \includegraphics[width=\linewidth]{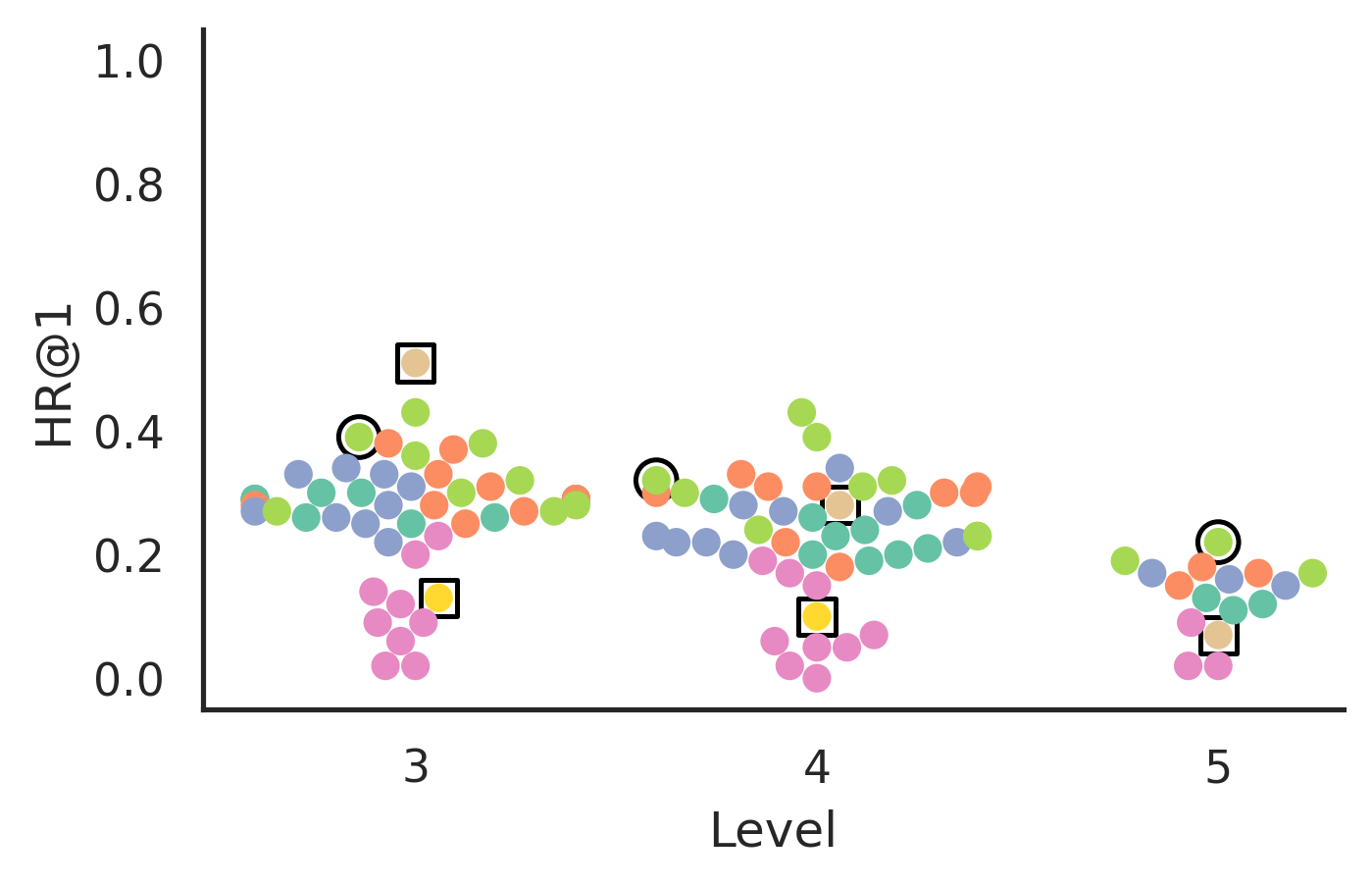}  
    \caption{HR@1.}
    \label{fig:real_hr1}
\end{subfigure}
\hfill
\begin{subfigure}{.49\textwidth}
   \centering
    \includegraphics[width=.7\linewidth]{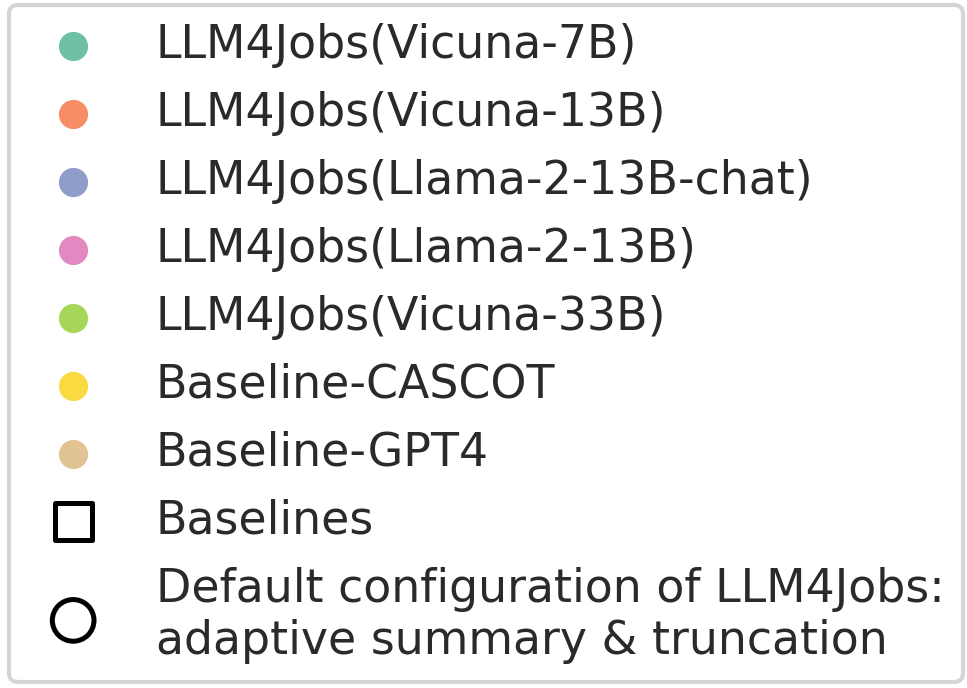}  
\end{subfigure}
\\
\begin{subfigure}{.49\textwidth}
    \centering
    \includegraphics[width=\linewidth]{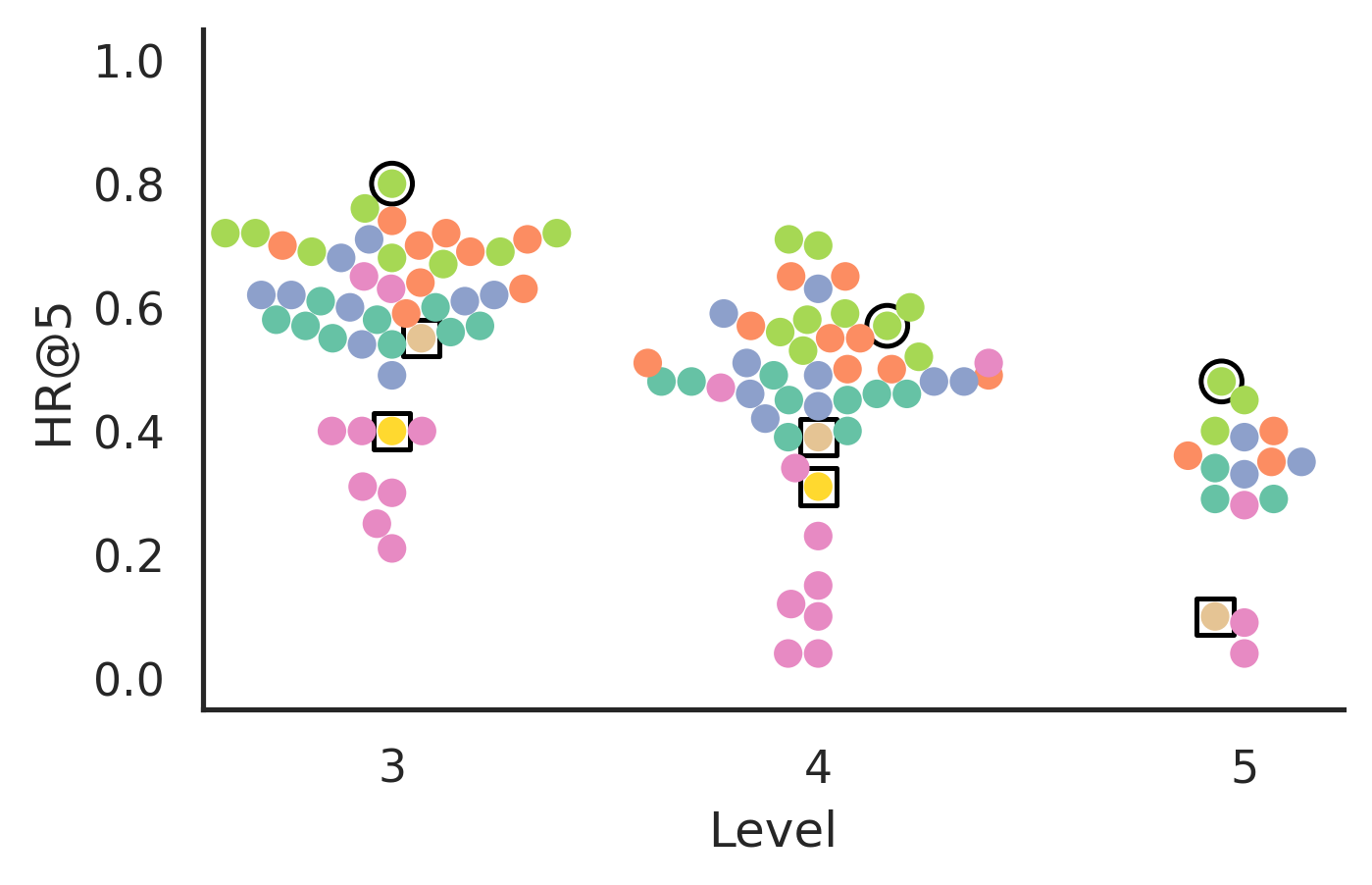}  
    \caption{HR@5.}
    \label{fig:real_hr5}
\end{subfigure}
\hfill
\begin{subfigure}{.49\textwidth}
    \centering
    \includegraphics[width=\linewidth]{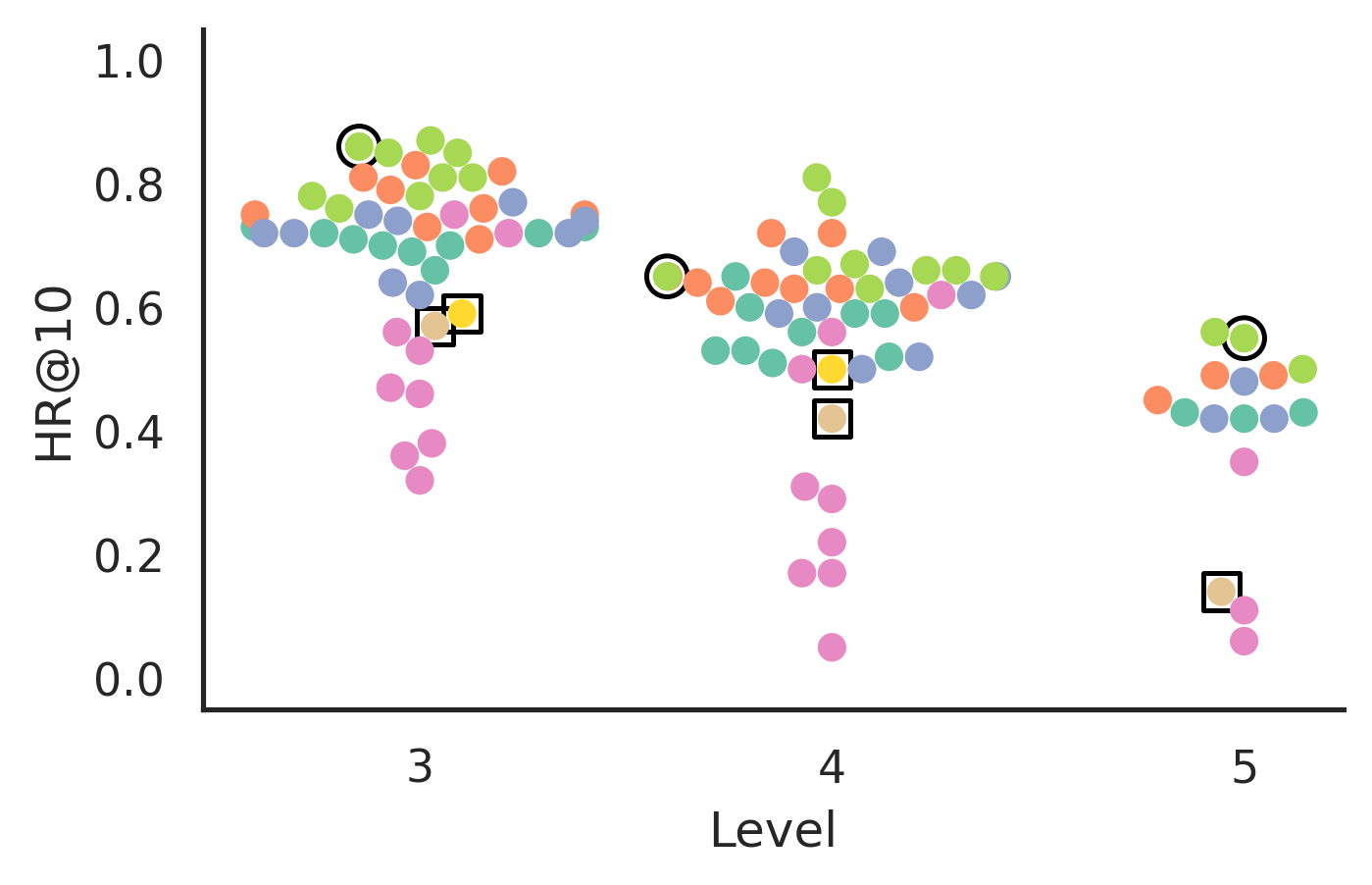}  
    \caption{HR@10.}
    \label{fig:real_hr10}
\end{subfigure}
\\
\begin{subfigure}{.49\textwidth}
   \centering
    \includegraphics[width=\linewidth]{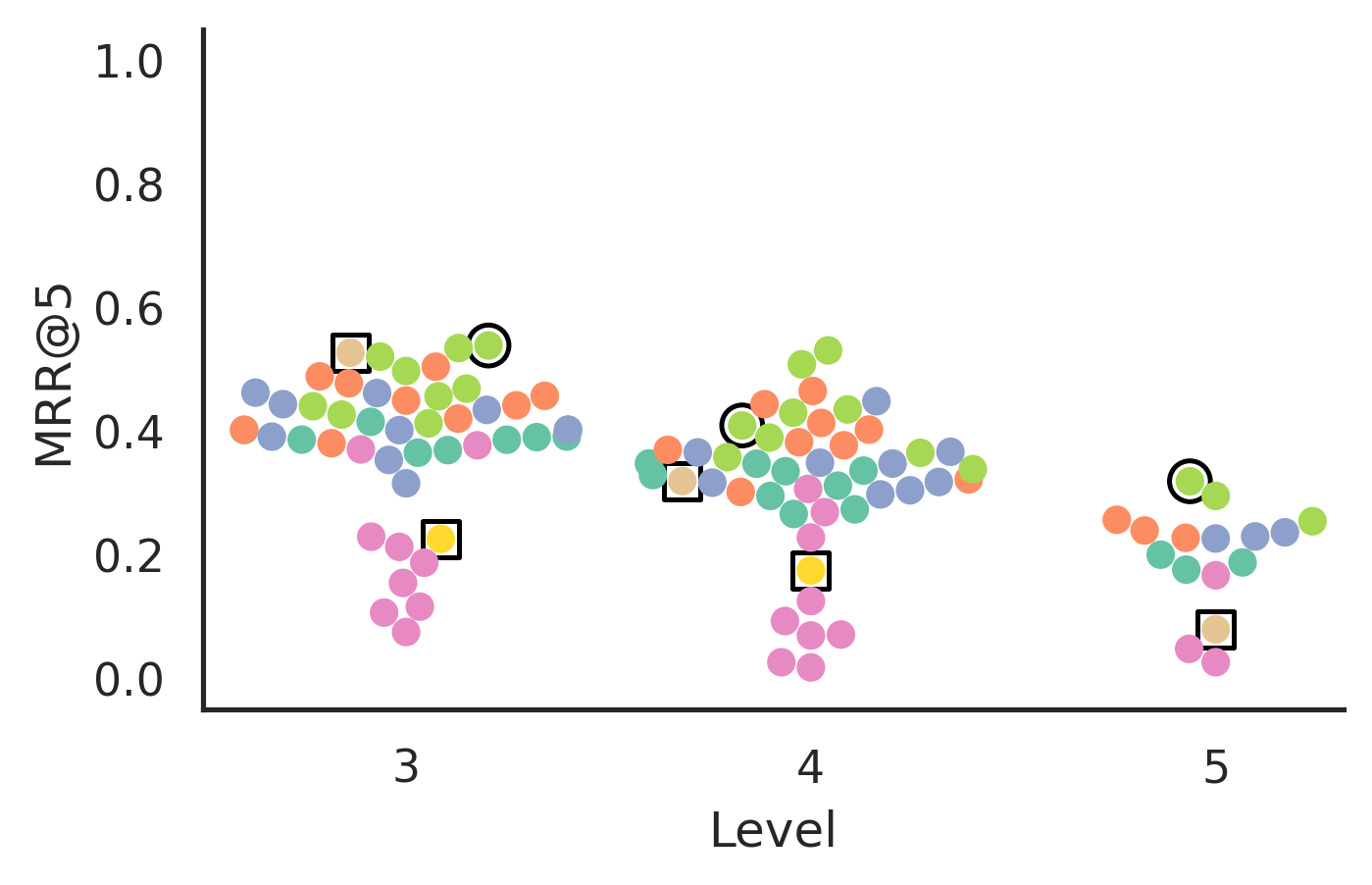}  
    \caption{MRR@5.}
    \label{fig:real_mrr5}
\end{subfigure}
\begin{subfigure}{.49\textwidth}
    \centering
    \includegraphics[width=\linewidth]{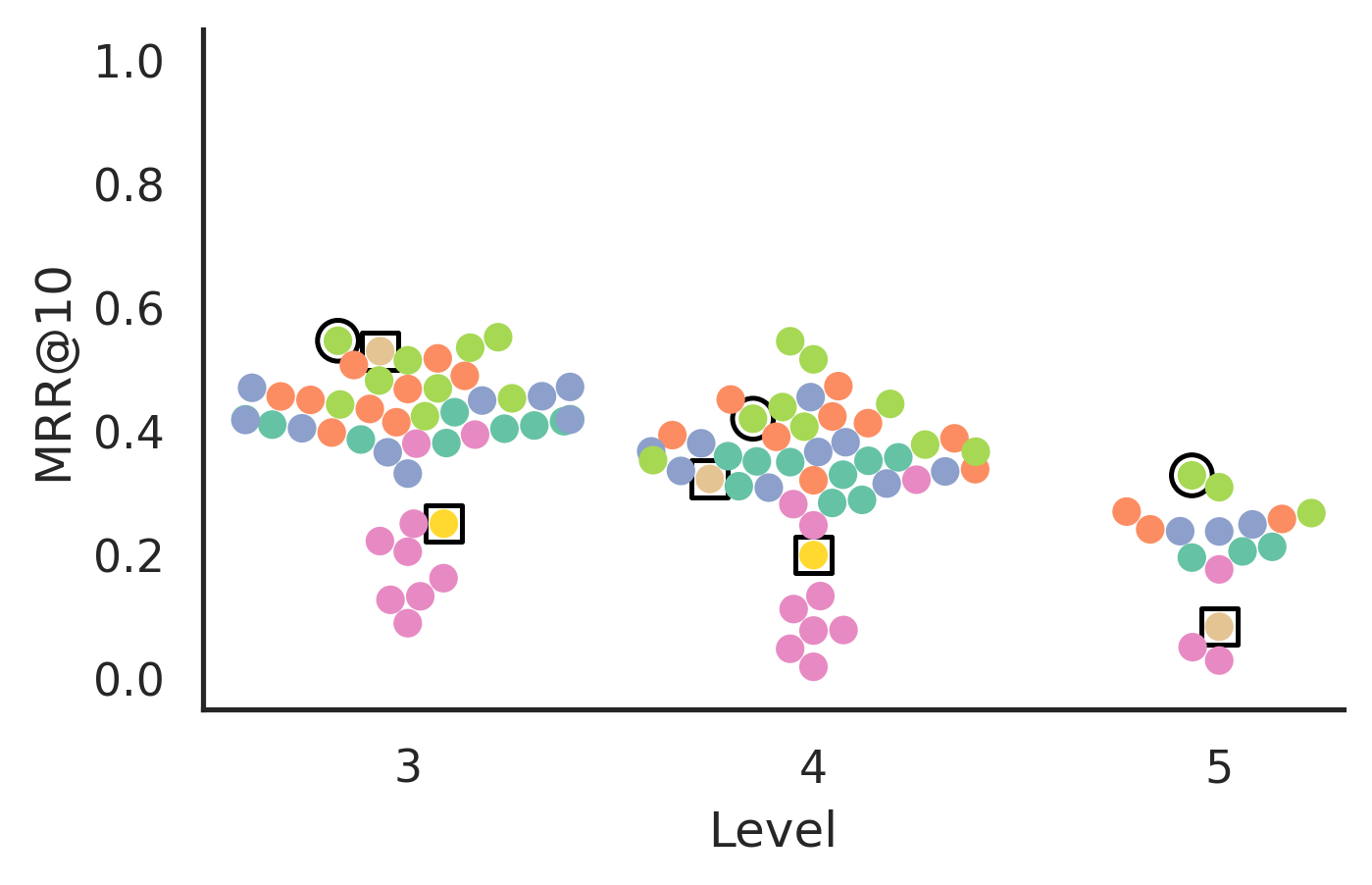}  
    \caption{MRR@10.}
    \label{fig:real_mrr10}
\end{subfigure}
\\
\begin{subfigure}{.49\textwidth}
    \centering
    \includegraphics[width=\linewidth]{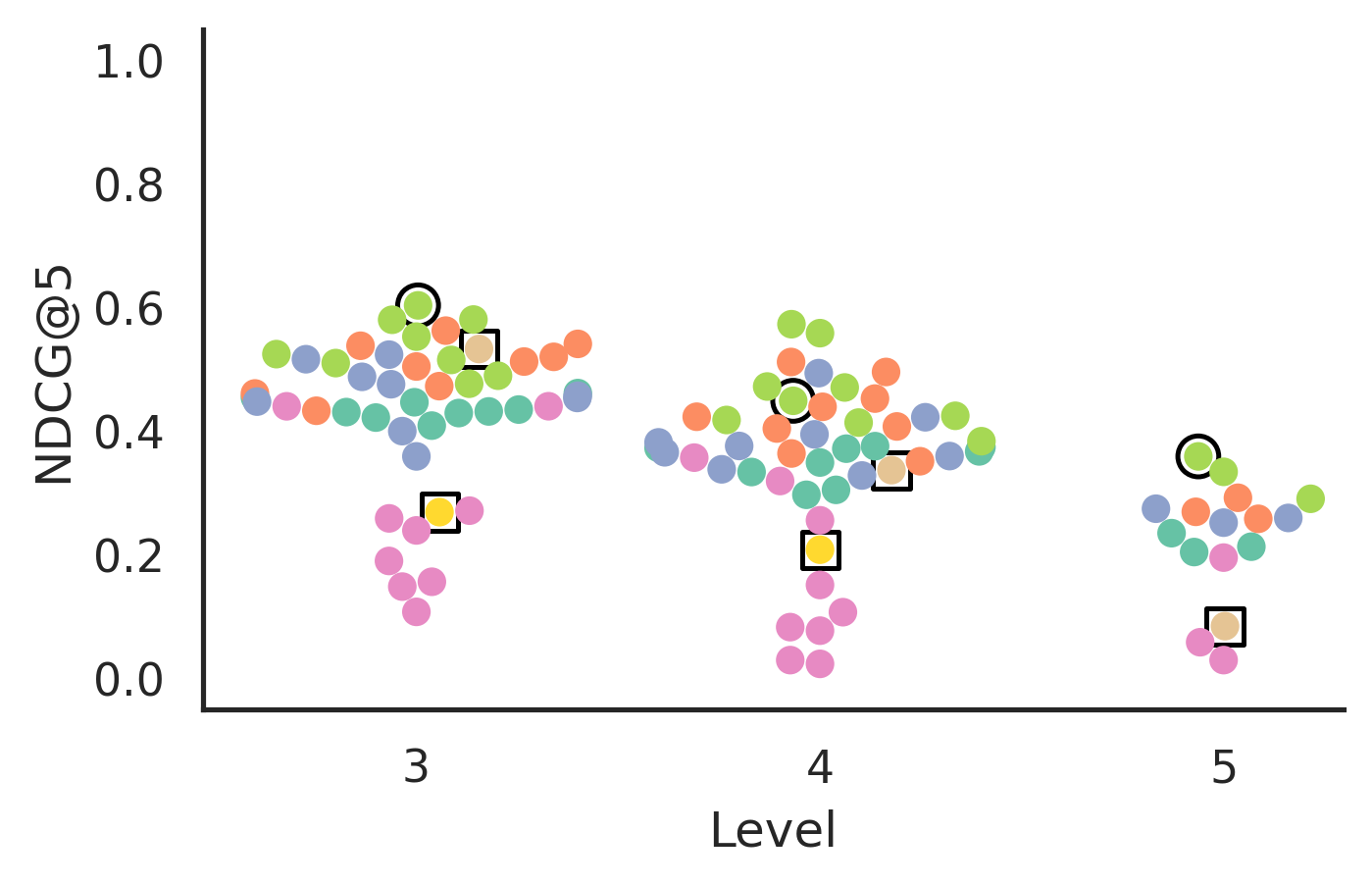}  
    \caption{NDCG@5.}
    \label{fig:real_ndcg5}
\end{subfigure}
\begin{subfigure}{.49\textwidth}
    \centering
    \includegraphics[width=\linewidth]{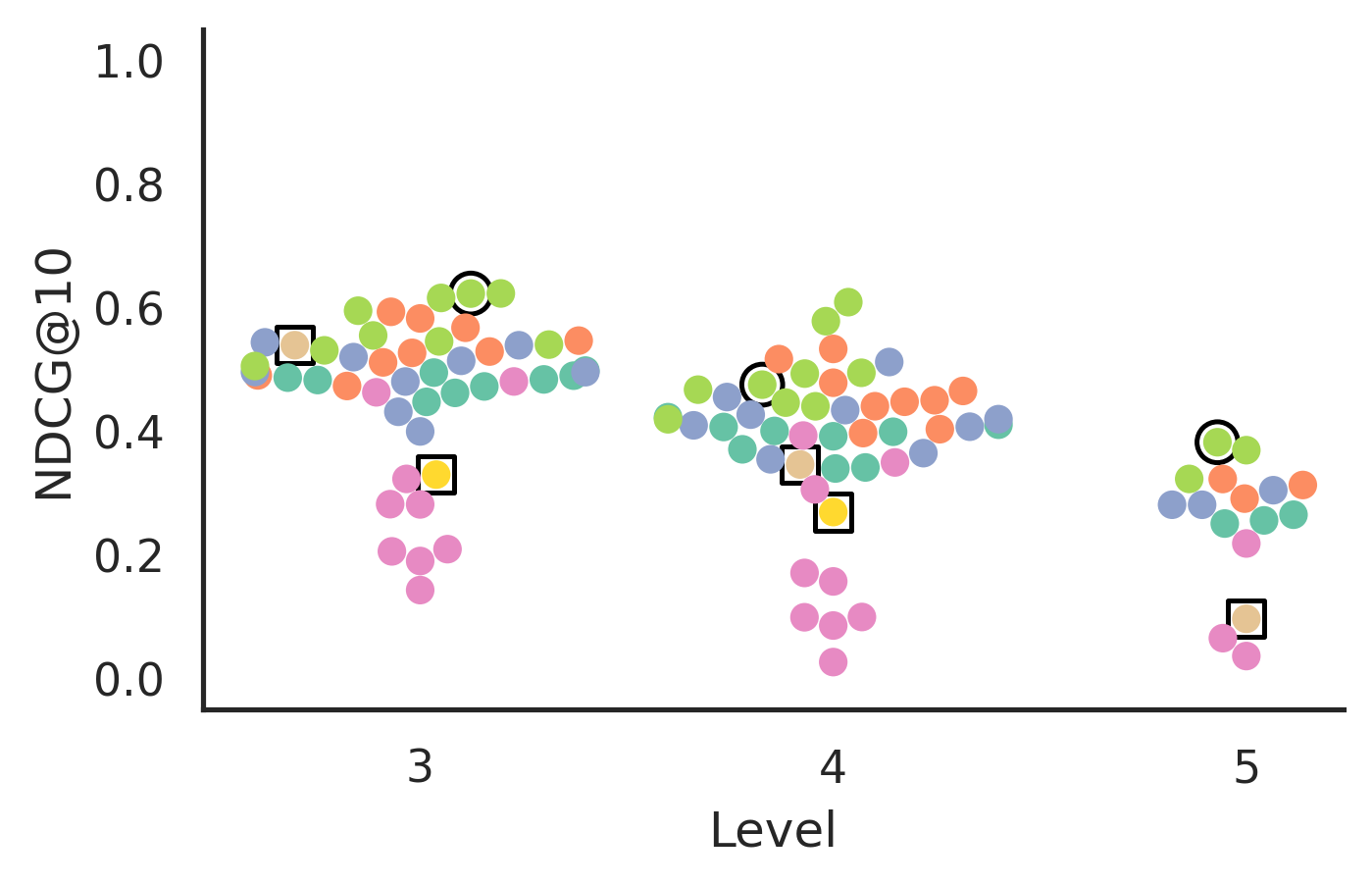}  
    \caption{NDCG@10.}
    \label{fig:real_ndcg10}
\end{subfigure}
\caption{All methods' performance on the \textbf{real-world} dataset. The points were jittered along the horizontal axis in order to minimize overlap.}
\label{fig:real_all}
\end{figure}

\begin{figure}
\raggedleft
\begin{subfigure}{.49\textwidth}
    \centering
    \includegraphics[width=\linewidth]{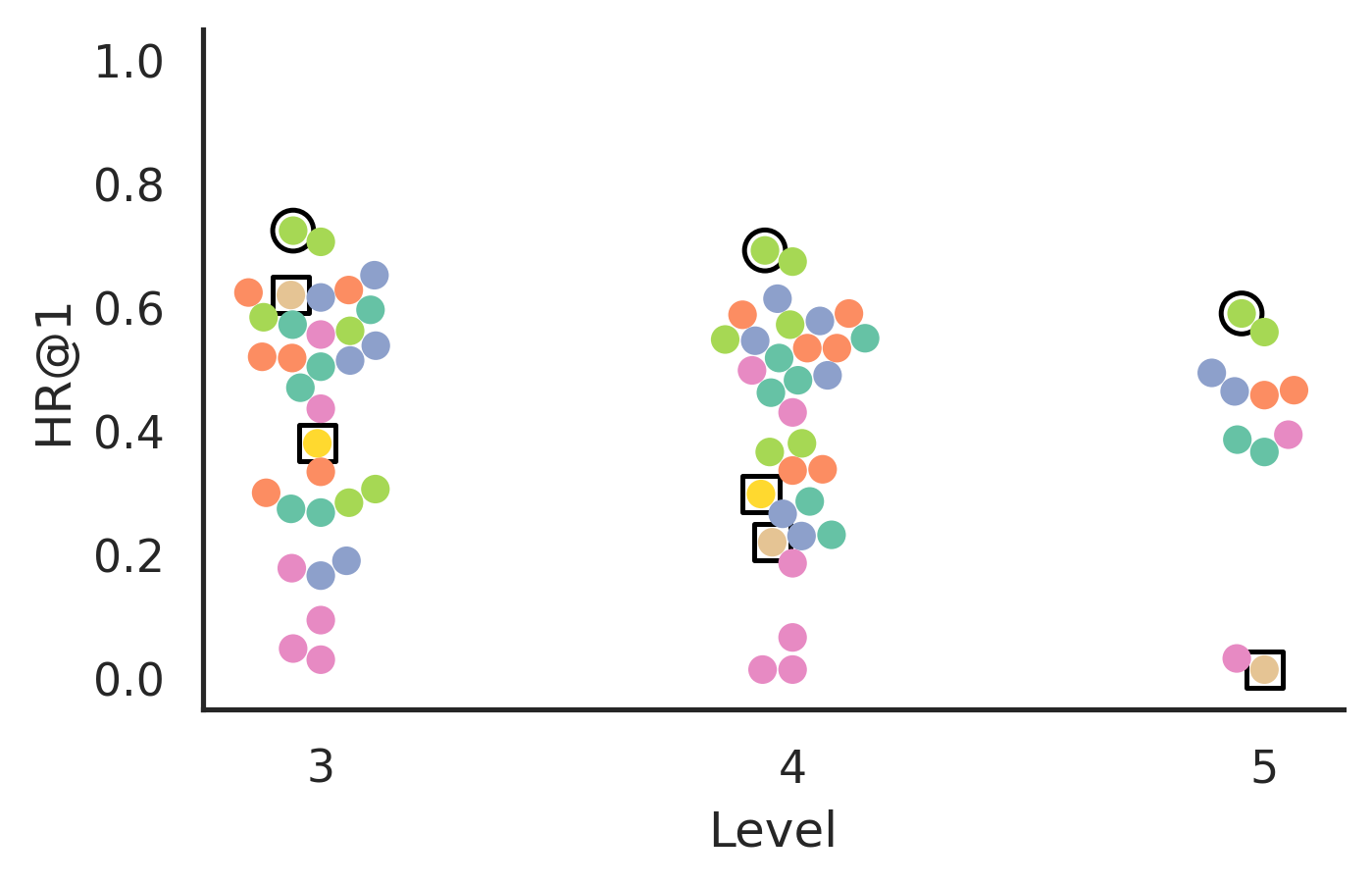}  
    \caption{HR@1.}
    \label{fig:easy_hr1}
\end{subfigure}
\hfill
\begin{subfigure}{.49\textwidth}
   \centering
    \includegraphics[width=.7\linewidth]{real_legend.png}  
\end{subfigure}
\\
\begin{subfigure}{.49\textwidth}
    \centering
    \includegraphics[width=\linewidth]{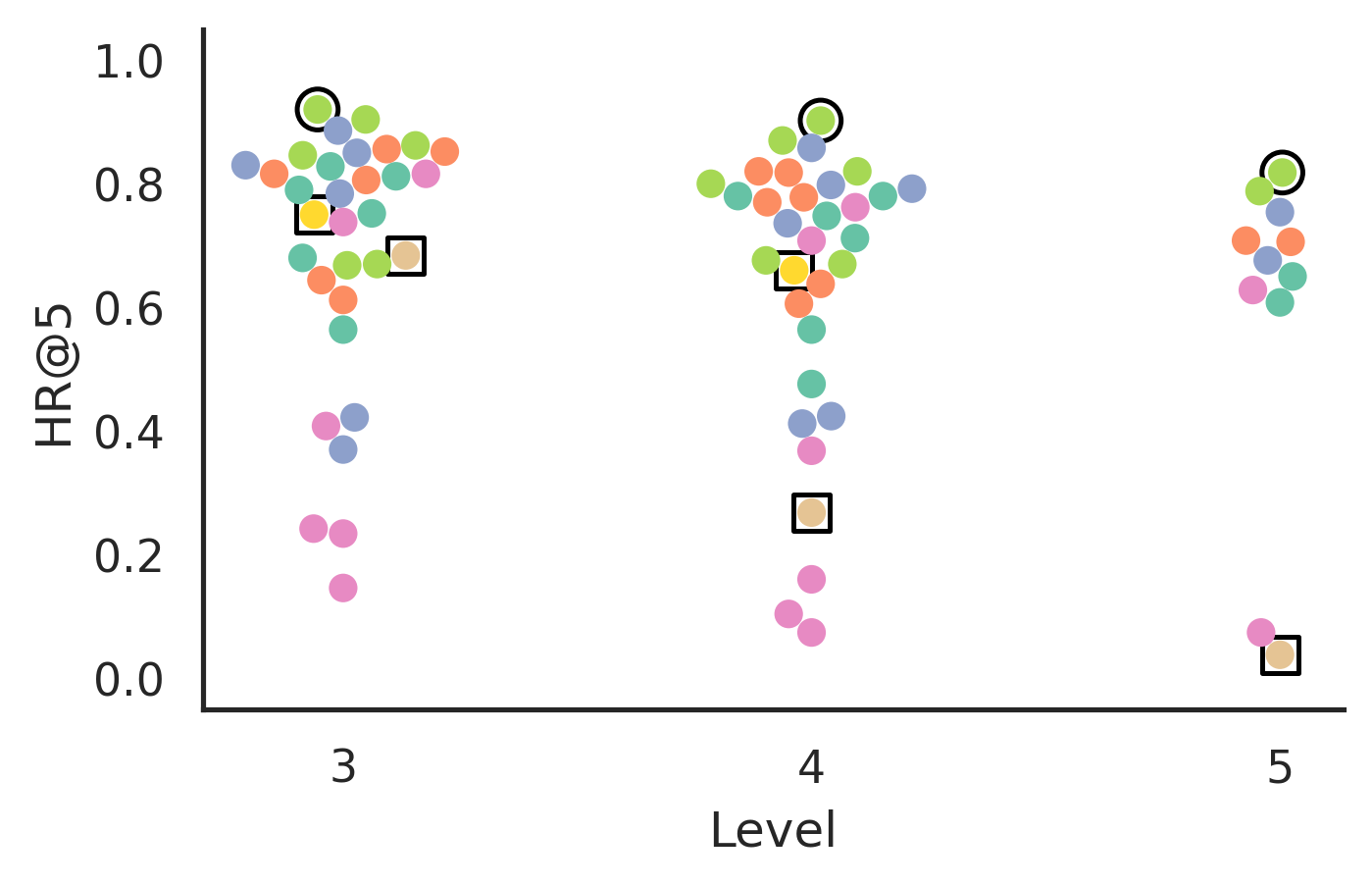}  
    \caption{HR@5.}
    \label{fig:easy_hr5}
\end{subfigure}
\hfill
\begin{subfigure}{.49\textwidth}
    \centering
    \includegraphics[width=\linewidth]{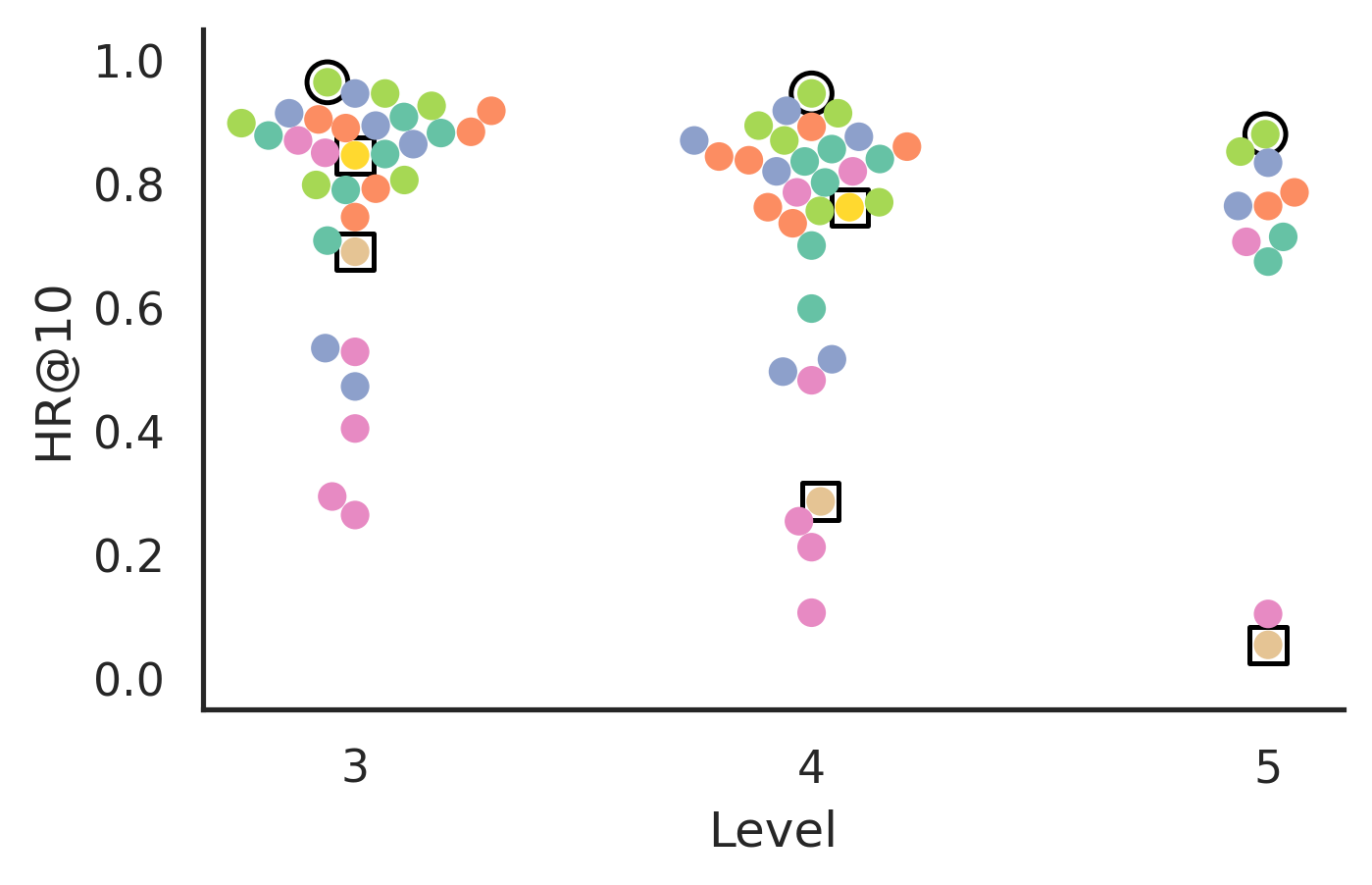}  
    \caption{HR@10.}
    \label{fig:easy_hr10}
\end{subfigure}\\

\begin{subfigure}{.49\textwidth}
   \centering
    \includegraphics[width=\linewidth]{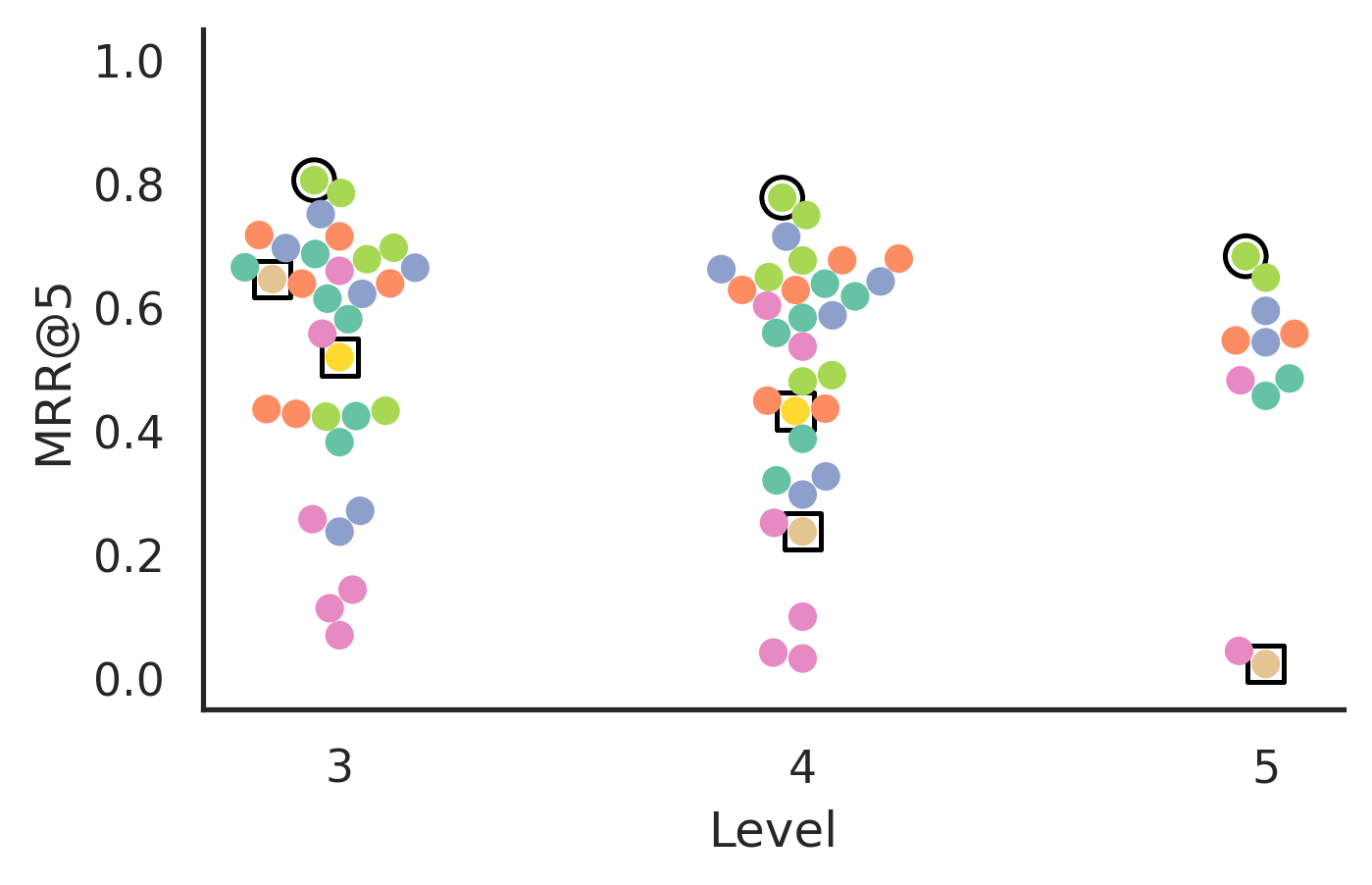}  
    \caption{MRR@5.}
    \label{fig:easy_mrr5}
\end{subfigure}
\begin{subfigure}{.49\textwidth}
    \centering
    \includegraphics[width=\linewidth]{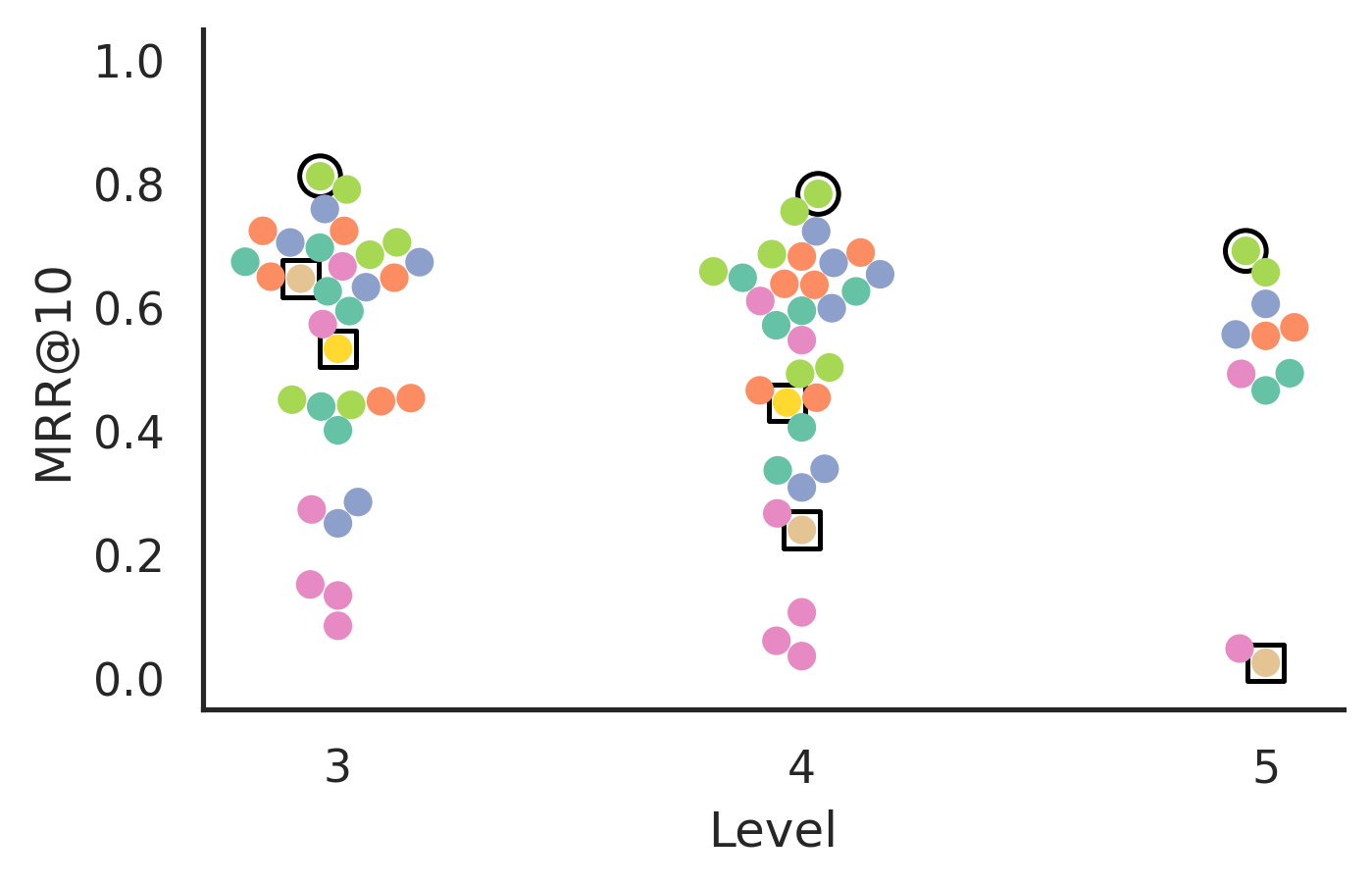}  
    \caption{MRR@10.}
    \label{fig:easy_mrr10}
\end{subfigure}
\\
\begin{subfigure}{.49\textwidth}
    \centering
    \includegraphics[width=\linewidth]{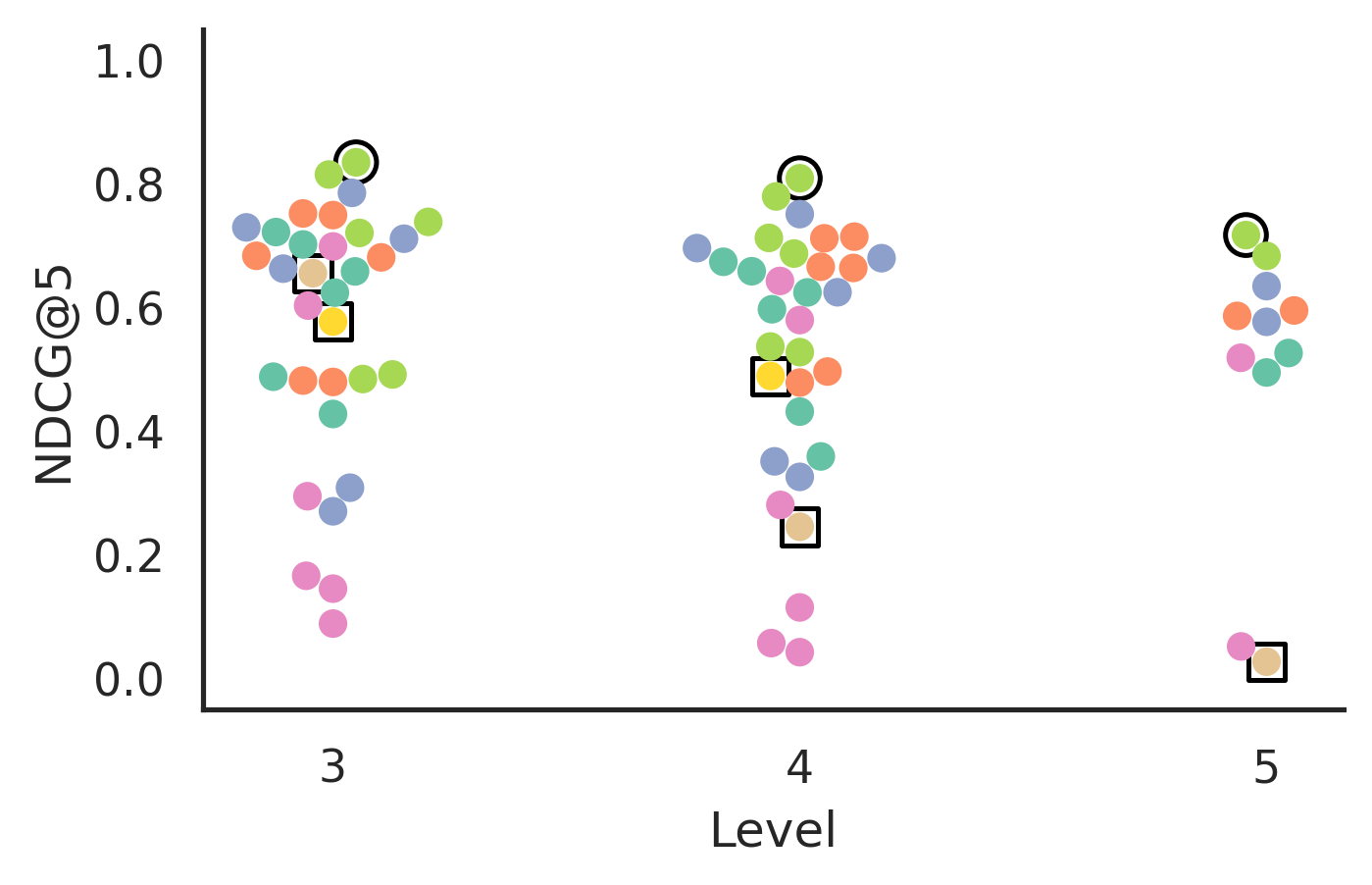}  
    \caption{NDCG@5.}
    \label{fig:easy_ndcg5}
\end{subfigure}
\begin{subfigure}{.49\textwidth}
    \centering
    \includegraphics[width=\linewidth]{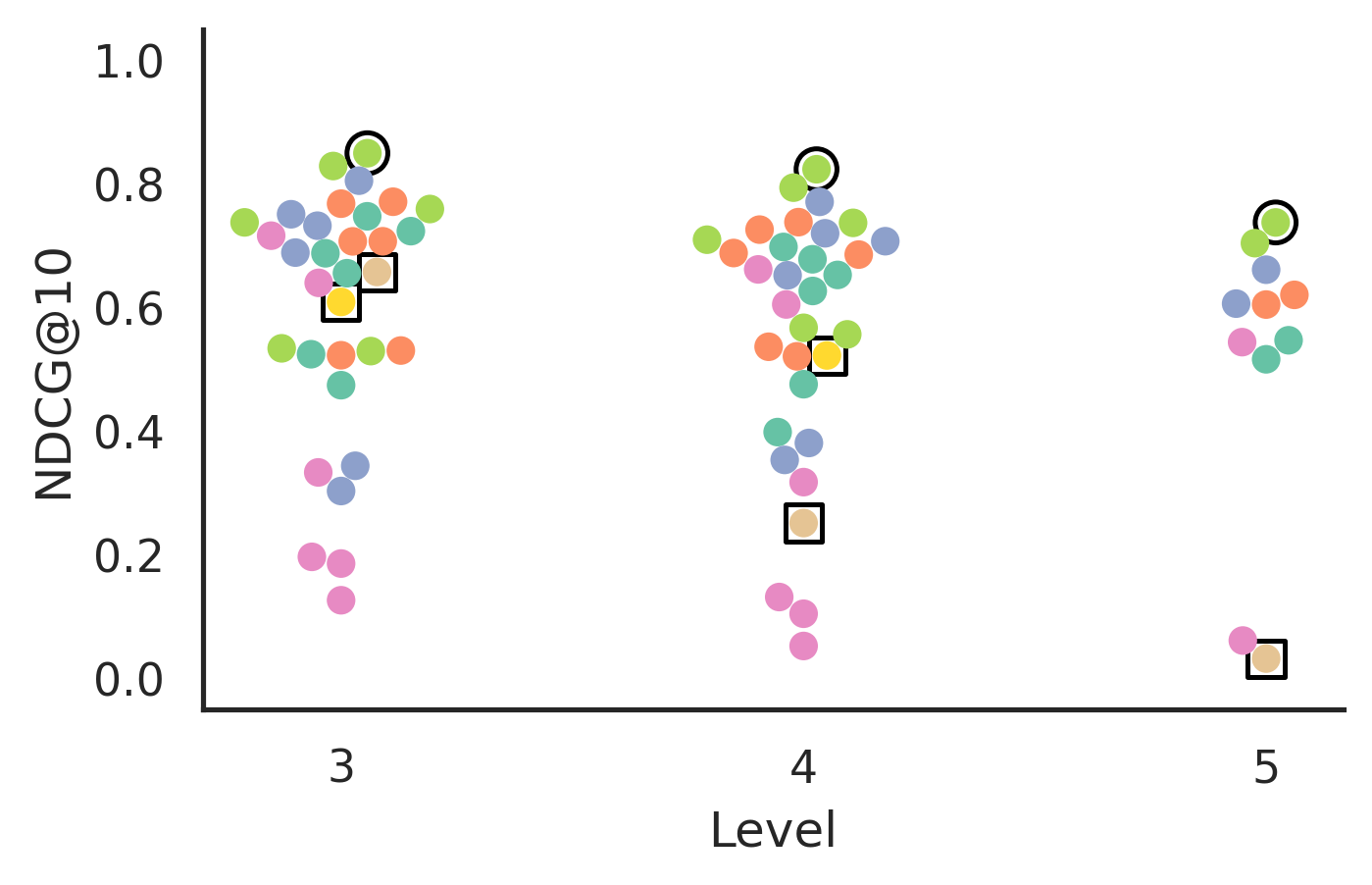}  
    \caption{NDCG@10.}
    \label{fig:easy_ndcg10}
\end{subfigure}
\caption{All methods' performance on the \textbf{GenEasy} dataset. The points were jittered along the horizontal axis in order to minimize overlap.}
\label{fig:easy_all}
\end{figure}

\begin{figure}
\raggedleft
\begin{subfigure}{.49\textwidth}
    \centering
    \includegraphics[width=\linewidth]{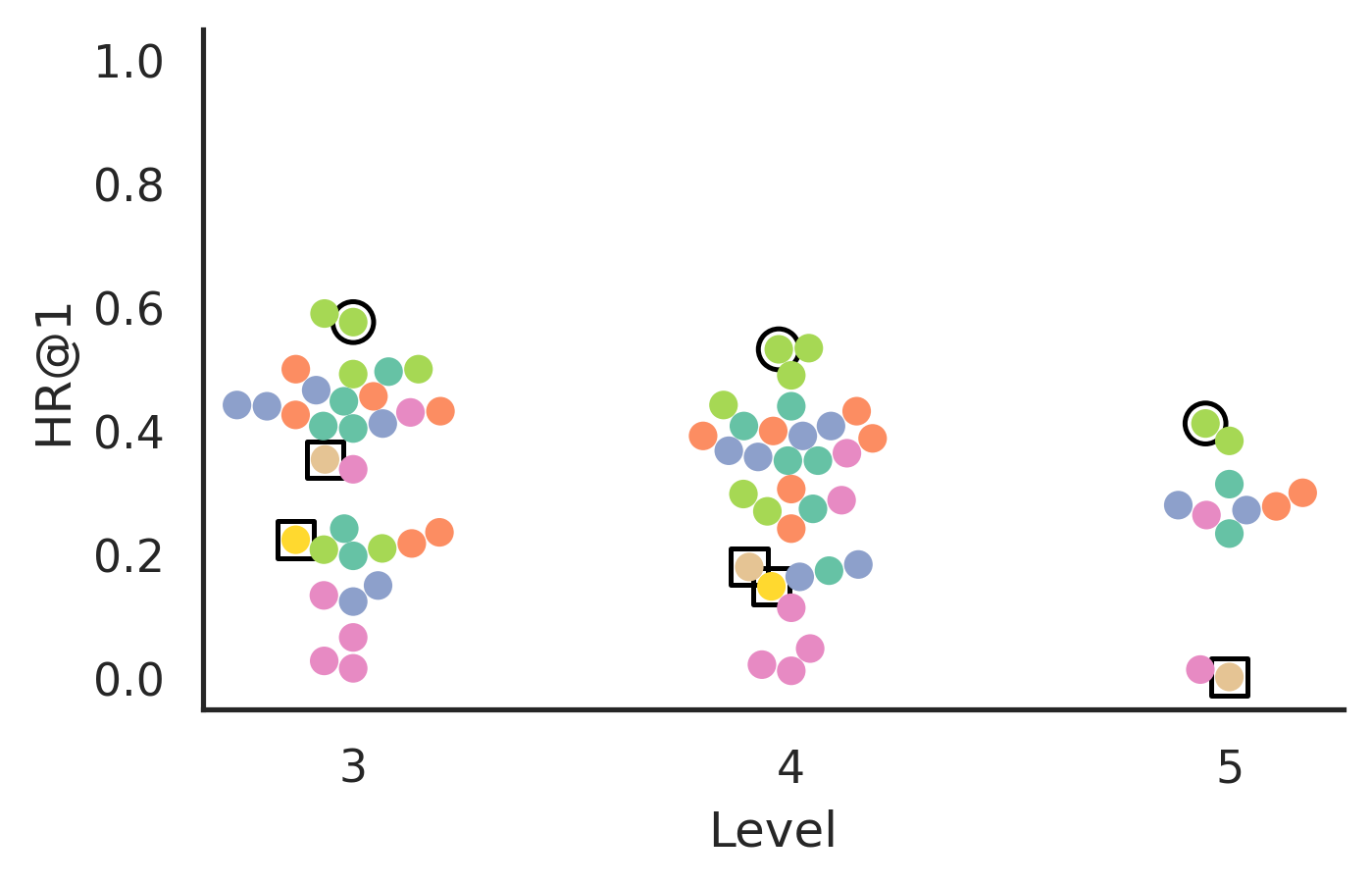}  
    \caption{HR@1.}
    \label{fig:hard_hr1}
\end{subfigure}
\hfill
\begin{subfigure}{.49\textwidth}
   \centering
    \includegraphics[width=.7\linewidth]{real_legend.png}  
\end{subfigure}
\\
\begin{subfigure}{.49\textwidth}
    \centering
    \includegraphics[width=\linewidth]{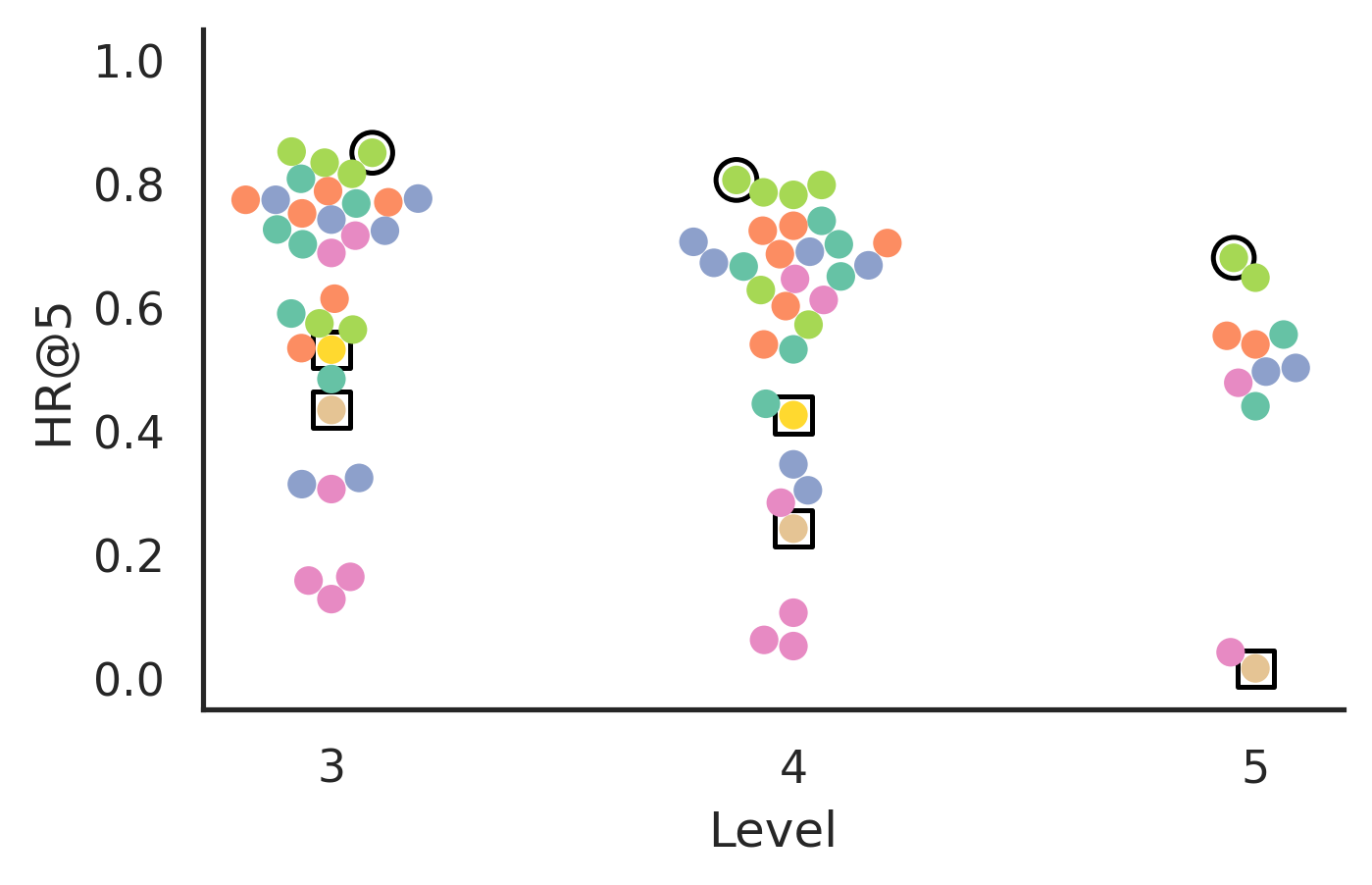}  
    \caption{HR5.}
    \label{fig:hard_hr5}
\end{subfigure}
\hfill
\begin{subfigure}{.49\textwidth}
    \centering
    \includegraphics[width=\linewidth]{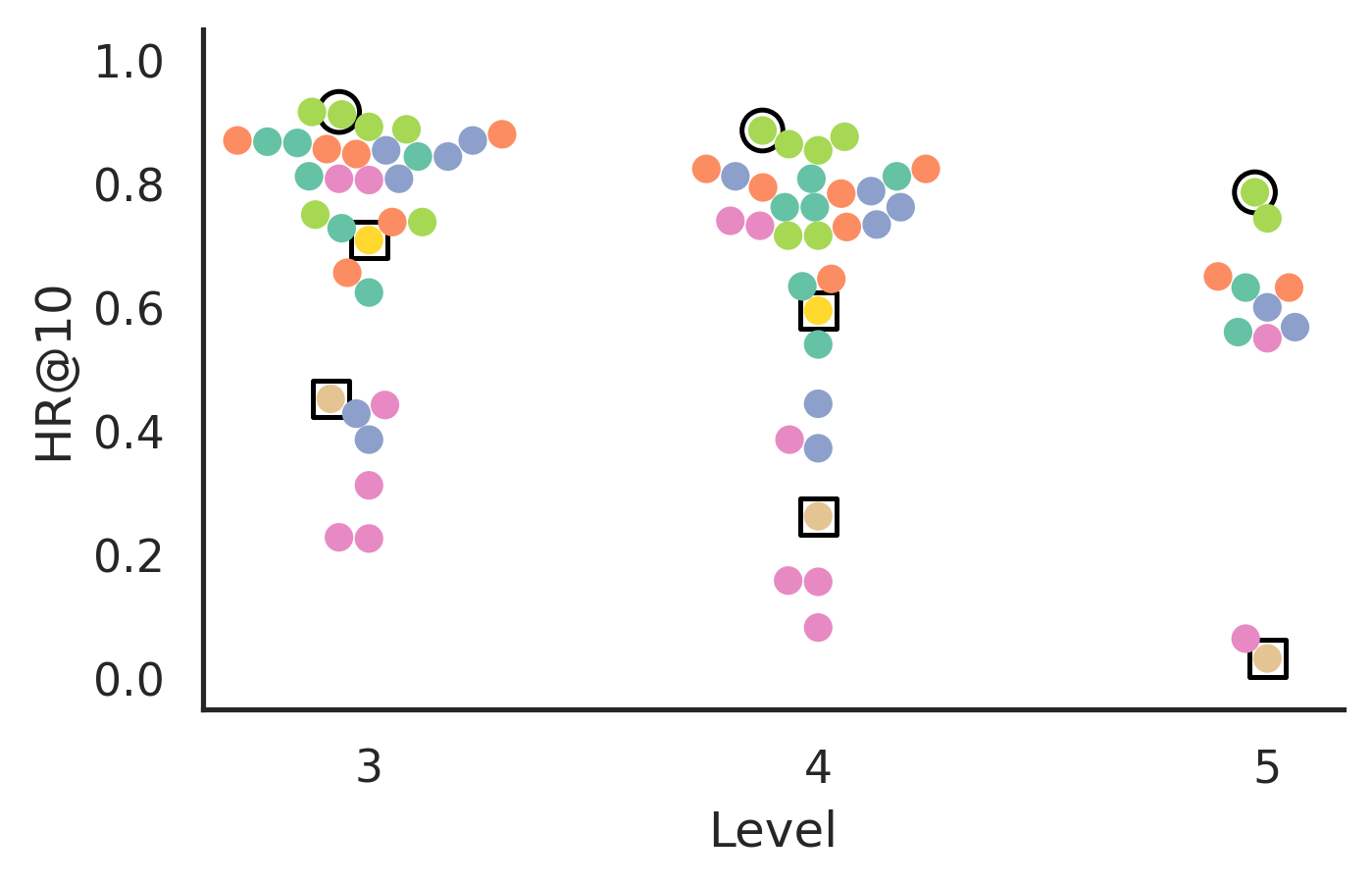}  
    \caption{HR@10.}
    \label{fig:hard_hr10}
\end{subfigure}
\\
\begin{subfigure}{.49\textwidth}
   \centering
    \includegraphics[width=\linewidth]{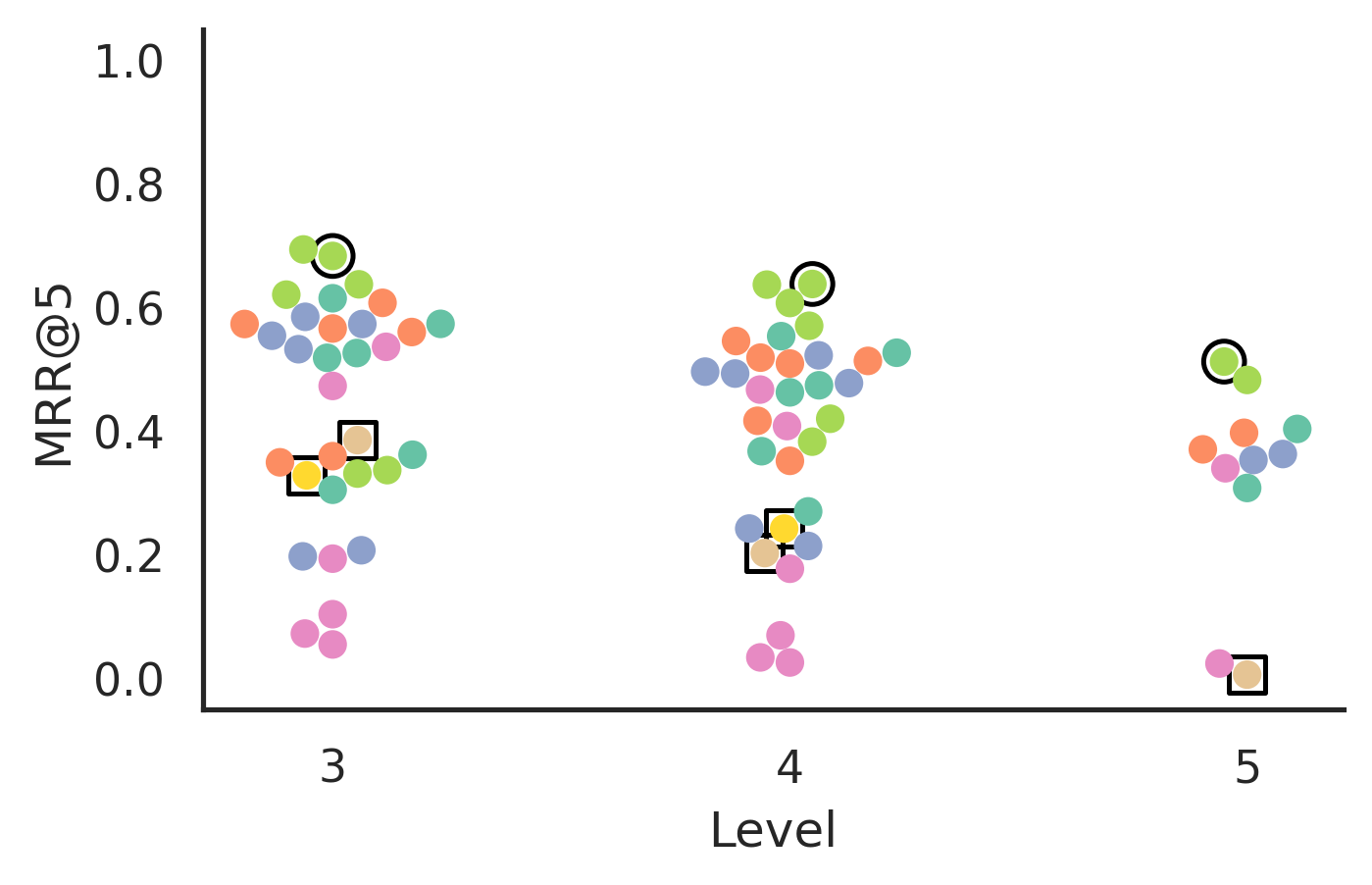}  
    \caption{MRR@5.}
    \label{fig:hard_mrr5}
\end{subfigure}
\begin{subfigure}{.49\textwidth}
    \centering
    \includegraphics[width=\linewidth]{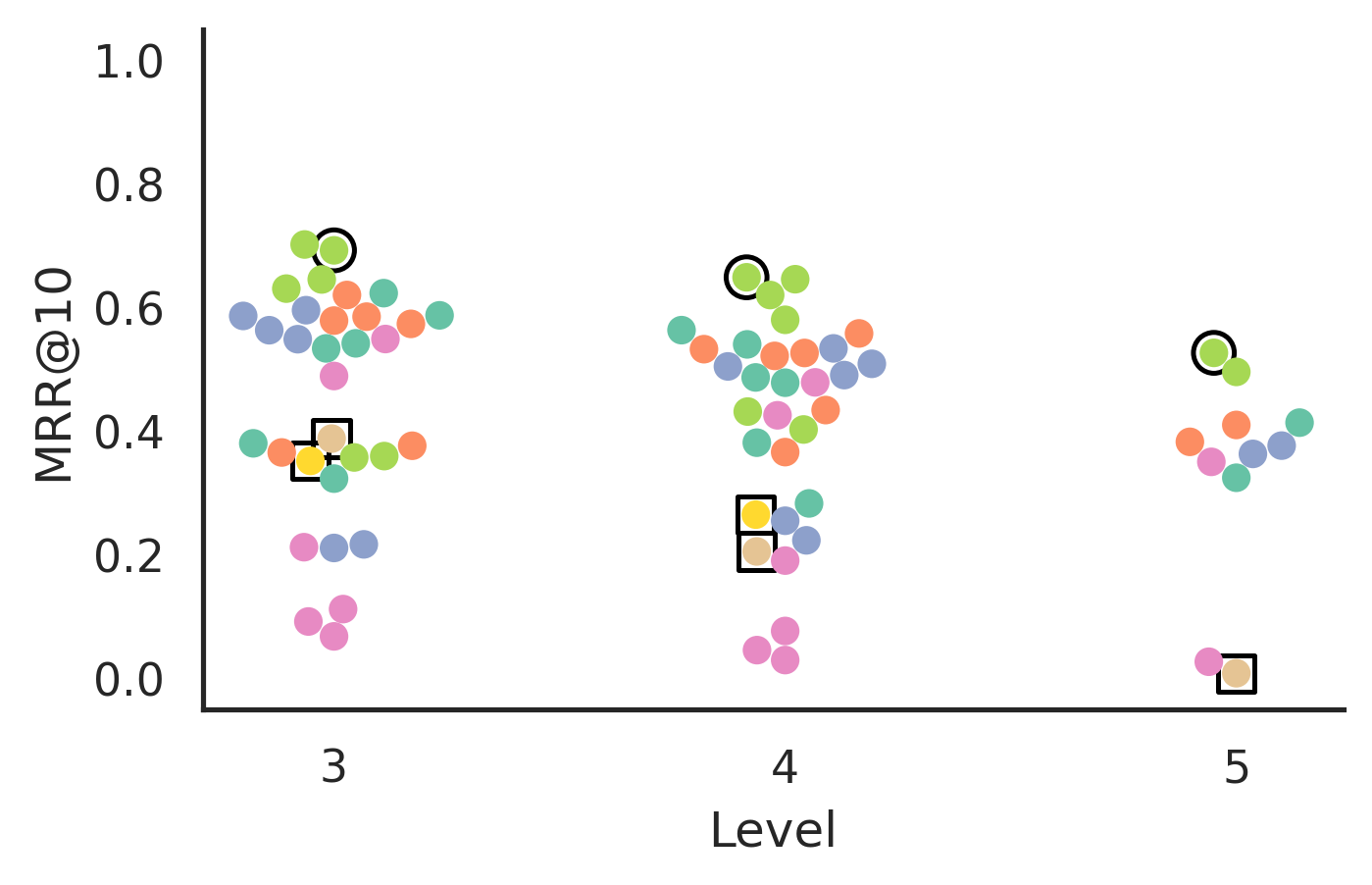}  
    \caption{MRR@10.}
    \label{fig:hard_mrr10}
\end{subfigure}
\\
\begin{subfigure}{.49\textwidth}
    \centering
    \includegraphics[width=\linewidth]{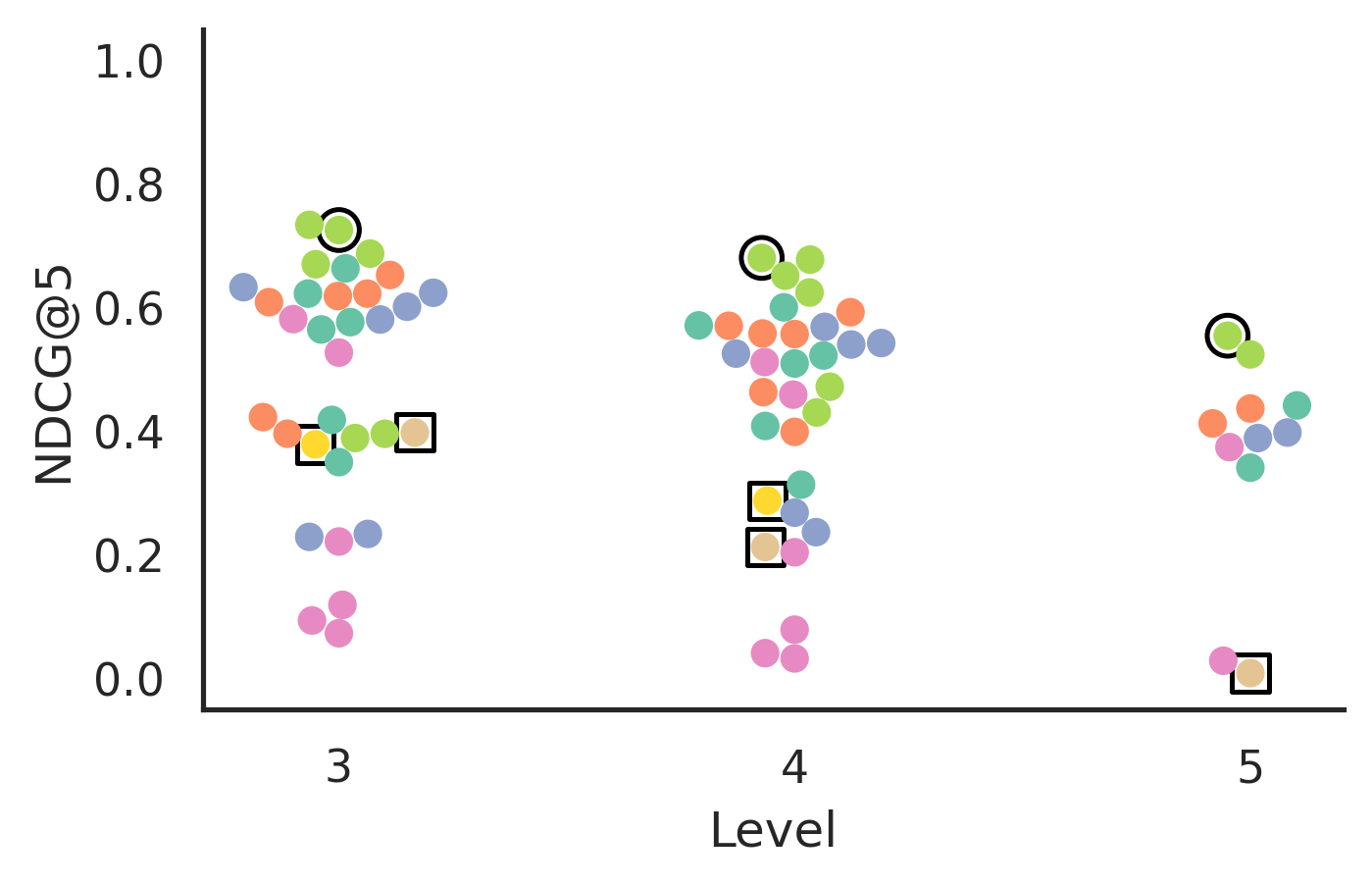}  
    \caption{NDCG@5.}
    \label{fig:hard_ndcg5}
\end{subfigure}
\begin{subfigure}{.49\textwidth}
    \centering
    \includegraphics[width=\linewidth]{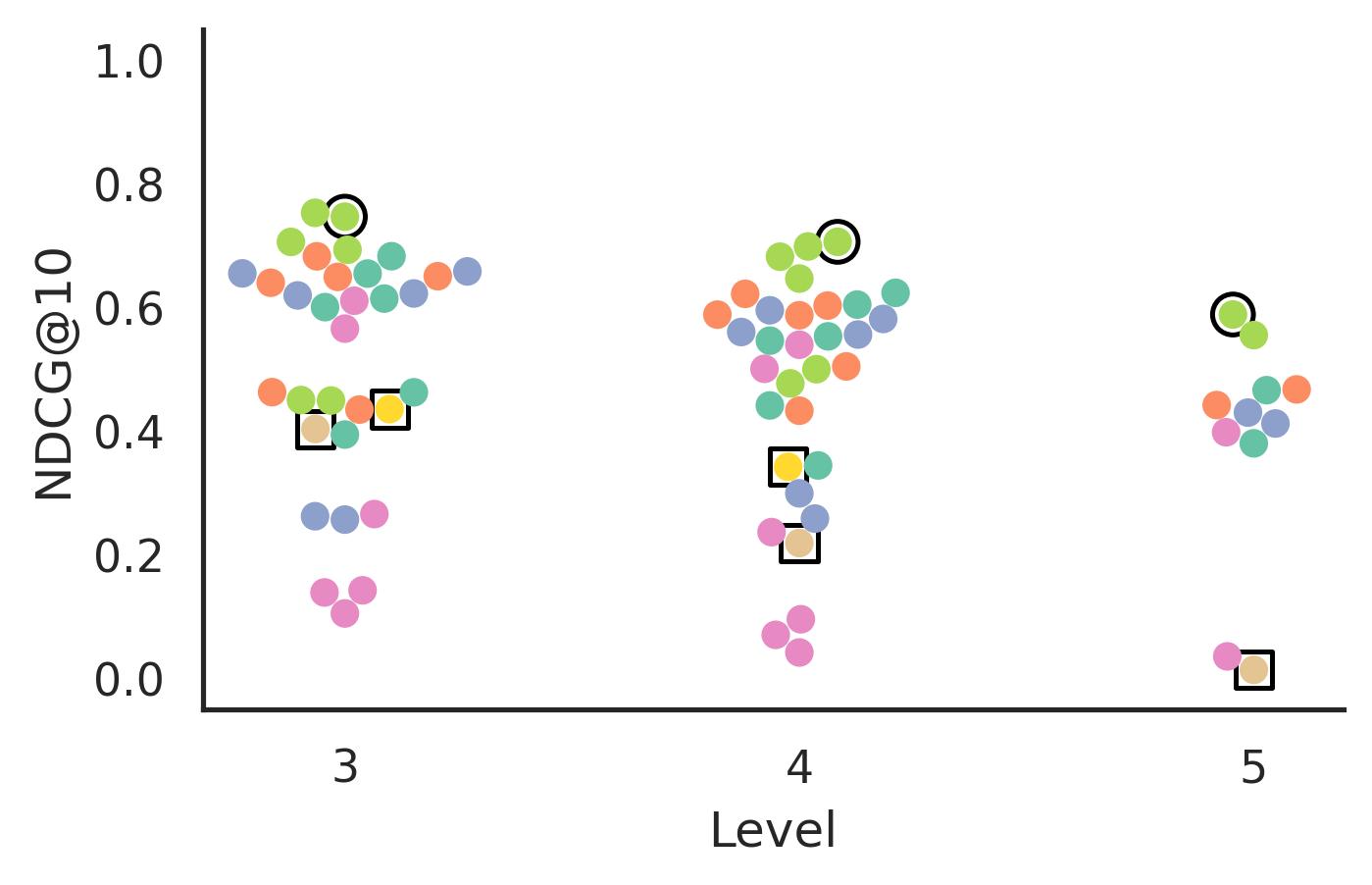}  
    \caption{NDCG@10.}
    \label{fig:hard_ndcg10}
\end{subfigure}
\caption{All methods' performance on the \textbf{GenHard} dataset. The points were jittered along the horizontal axis in order to minimize overlap.}
\label{fig:hard_all}
\end{figure}

Performance across various configurations of LLM4Jobs with different backbone models in comparison to the baselines for the three datasets is visualized in Fig.~\ref{fig:real_all}, \ref{fig:easy_all}, and \ref{fig:hard_all} for RQ\ref{rq3}.

\subsection{Performance of LLM4Jobs vs. Baselines (RQ\ref{rq1})}\label{sec:rq1_results}

The following observations can be drawn regarding LLM4Jobs in comparison to the baselines:

\textbf{Consistent Superiority of LLM4Jobs}: Across all datasets and granularities, LLM4Jobs consistently surpasses both CASCOT and GPT4 in nearly all metrics.

\textbf{Performance Gradient by Dataset}: All methods generally perform best on the GenEasy dataset, followed closely by the GenHard dataset. The real-world dataset, as expected, poses the most challenges. 

\textbf{Granularity Challenges}: Across all methods, as we move from Level 3 to Level 5+ granularity, there's a noticeable drop in performance. This aligns with the intuition that finer granularities introduce more complexity due to the increased specificity required.

\textbf{Real-world dataset nuances}: On the real-world dataset (Table \ref{tab:real}), LLM4Jobs exhibits a clear advantage, especially at Level 5 and 4 granularities. It handles the inherent complexities and nuances present in genuine job postings more effectively than the baselines.

\textbf{Performance on Synthetic Datasets}: On the GenHard and GenEasy datasets (Tables \ref{tab:easy} and \ref{tab:hard}), LLM4Jobs consistently outperforms the baselines, but the margin of superiority varies. On the GenEasy dataset, the performance gaps are narrower, suggesting that when the data is clearer, the baselines can achieve results closer to LLM4Jobs. However, the more challenging GenHard dataset widens the performance gaps, emphasizing LLM4Jobs' capability to handle nuanced and intricate job descriptions.

\textbf{GPT4's Dynamics}: GPT4's performance, especially in the real-world dataset, might be impacted by the hallucination problem, where it suggests non-existent codes, especially for the level-5+ codes. This is a known limitation of LLMs and might explain GPT4's relatively lower performance. 
However, its top performance on HR@1 for Level 3 granularity indicates its ability to identify broader occupation categories. 

\textbf{CASCOT's Limitations}: CASCOT, despite being a state-of-the-art tool for ISCO coding, seems to struggle, especially at finer granularities. This underlines the challenges of rule-based systems in the face of diverse and sometimes ambiguous job descriptions.

In summary, while the advantage of LLM4Jobs over the baselines is evident across all datasets, the degree of this advantage varies based on the dataset's complexity and granularity level. The results establish LLM4Jobs' position as a leading method for unsupervised occupation extraction and standardization.

\subsection{Variations of LLM4Jobs (RQ\ref{rq2})}\label{sec:rq2_results}
Here we discuss the relative performance of different configurations of LLM4Jobs.

\textbf{Real-world dataset}: On this dataset (Table~\ref{tab:real}), any form of summarization—whether applied to all documents or adaptively based on length—outperforms the no-summarization approach. The results suggest that extracting occupation-centric information from real job postings enhances categorization.

Different granularities show varied preferences for mapping strategies. At level 4, direct mapping is often superior, whereas at level 3, truncation occasionally takes the lead. Clustering consistently lags.

\textbf{Synthetic datasets}: On these datasets (Table~\ref{tab:easy}, \ref{tab:hard}), configurations without summarization often match or surpass the ``all summary'' approach at Levels 4 and 5. This might indicate that the inherently clear nature of synthetic data renders additional summarization redundant.

For mapping strategy, truncation frequently emerges as the preferred method, indicating that accurate predictions at finer granularities can simply be truncated to yield accurate higher-level predictions.

\subsection{Influence of Backbone LLMs (RQ\ref{rq3})}\label{sec:rq3_results}
Here we study the effect of using more or less capable LLMs.

\textbf{Model size impact}: As expected, the largest model Vicuna-33b tend to outperform its smaller counterparts, consistent with the results from other benchmarks (\cite{zheng2023judging}). Nonetheless, smaller models such as Vicuna-13b and LLaMA-2-13b-chat sometimes have comparable performances (e.g. Fig.~\ref{fig:real_hr1}, \ref{fig:real_hr5}, \ref{fig:real_hr10}). 

\textbf{Human alignment impact}: Different model versions, even with comparable sizes, exhibit varied performance. Among the 13-billion parameter contenders, Vicuna stands out, affirming its efficacy as reported in other studies. Similarly, LLaMA-2-chat, fine-tuned for dialogue use cases, surpasses LLaMA-2. In fact, LLaMA-2, the basic language model without any human alignment, yielded the worst performance, often even worse than the baselines (e.g. Fig.~\ref{fig:easy_hr5}, \ref{fig:hard_ndcg5}). These results underscore the importance of human alignment.

\textbf{Trade-offs}: While larger models like Vicuna-33b offer superior performance, there's a trade-off in terms of computational resources and inference time. Deploying such models in real-world, especially time-sensitive scenarios, would require careful consideration of these trade-offs.

\subsection{Qualitative Analysis}
To gain additional insight into the behavior, strengths and limitations of LLM4Jobs, and thus fruitful avenues for further work, we conducted a number of qualitative analyses.
\subsubsection{Quality of taxonomy embeddings}

\begin{figure}[!h]
\includegraphics[width=\textwidth]{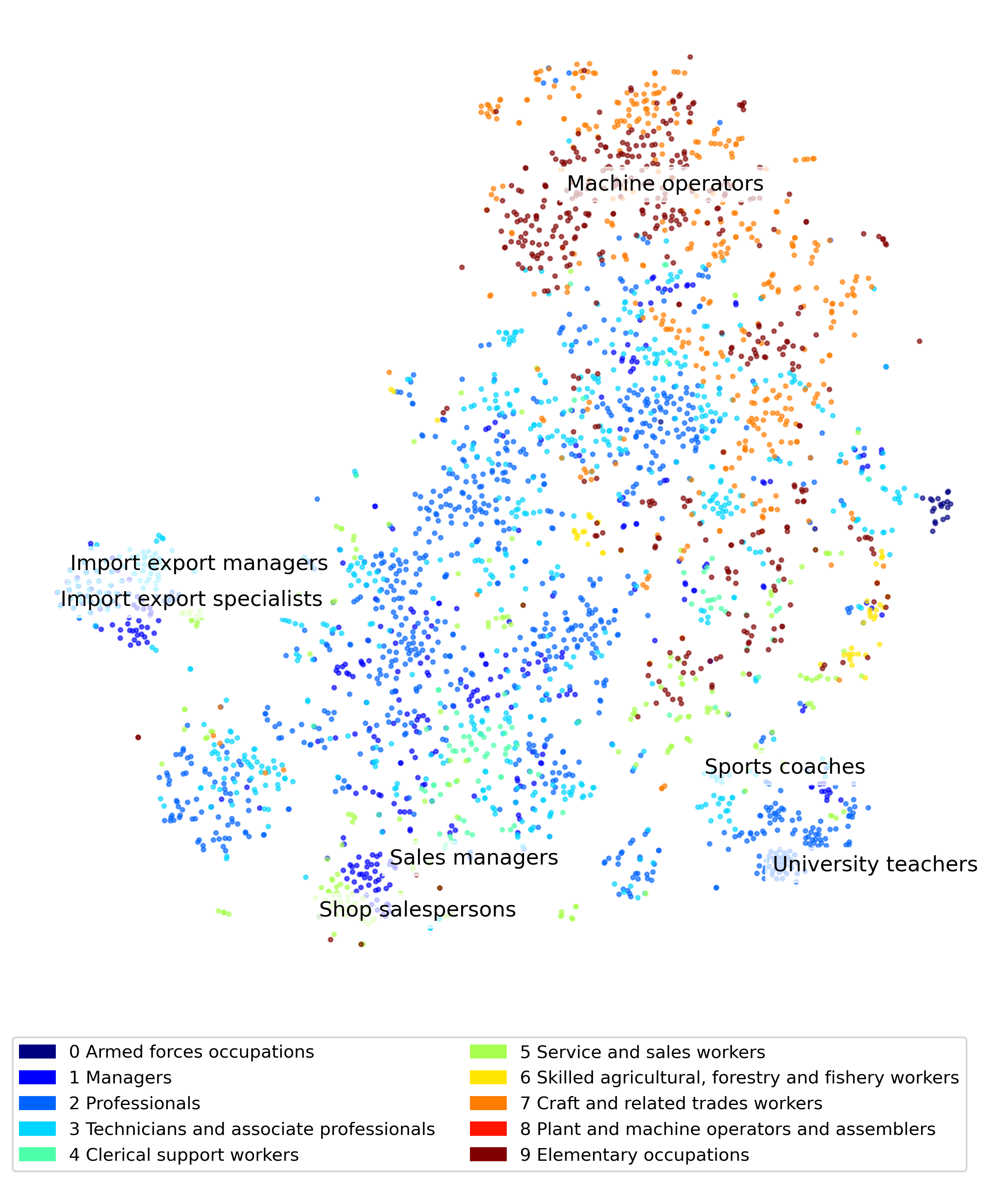}
\caption{Visualization of the taxonomy embeddings using t-SNE, generated by Vicuna-33b.}
\label{fig:tsne_embedding}
\end{figure} 

The t-SNE visualization of the taxonomy embeddings produced by Vicuna-33b (as shown in Figure~\ref{fig:tsne_embedding}) highlights the clustering of occupations based on their semantic similarities. While this clustering often aligns with the ESCO taxonomy, there are instances where the semantic closeness leads to intermixing of occupations that fall under different high-level codes.

For instance, while occupations related to ``university and higher education teachers'' and ``machine operators'' form distinct clusters, the occupations ``Import export managers'' and ``Import export specialists'' are closely positioned, despite belonging to separate top-level codes (1324.3.2 and 3331.2.1, respectively). Similarly, ``Sales managers'' and Shop salespersons'' are closely aligned due to shared semantic aspects, even though they belong to codes 1 and 5, respectively.

Such overlaps can challenge LLM4Jobs' ability to distinguish between high-level codes, as the semantic nuances can sometimes overshadow the strict hierarchical distinctions. A backbone model specifically fine-tuned for the occupation taxonomy could potentially offer more refined embeddings that better respect the taxonomy's structure. Such efforts present promising avenues for future research.

\subsubsection{Effect of summarization}
The efficacy of the summarization step in enhancing accuracy, especially for genuine job postings, is clear. By examining specific cases, we can better understand its impact.

\begin{figure}[!h]
\includegraphics[width=0.8\textwidth]{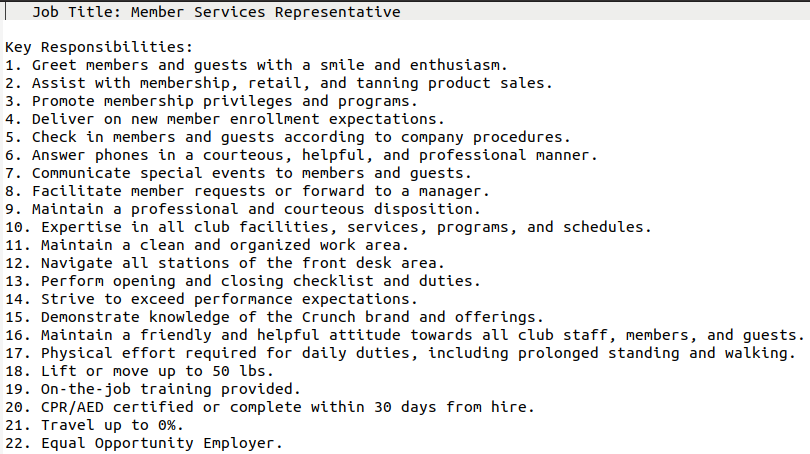}
\caption{Summary for a job posting titled ``Member Service Representative''.}
\label{fig:member_summary}
\end{figure} 

\begin{figure}[!h]
\includegraphics[width=\textwidth]{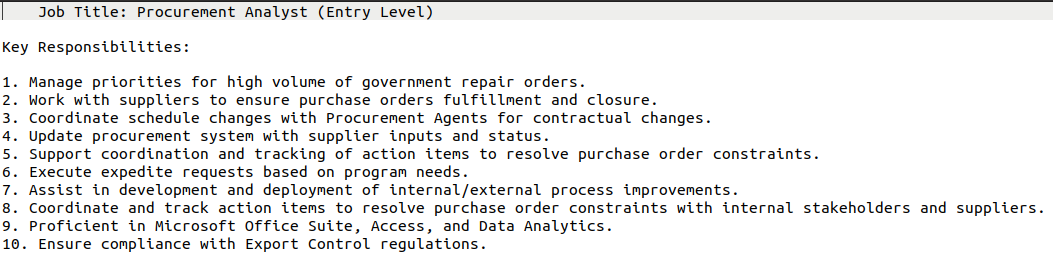}
\caption{Summary for a job posting titled ``Procurement Analyst''.}
\label{fig:procurement_summary}
\end{figure} 

\begin{figure}[!h]
\includegraphics[width=\textwidth]{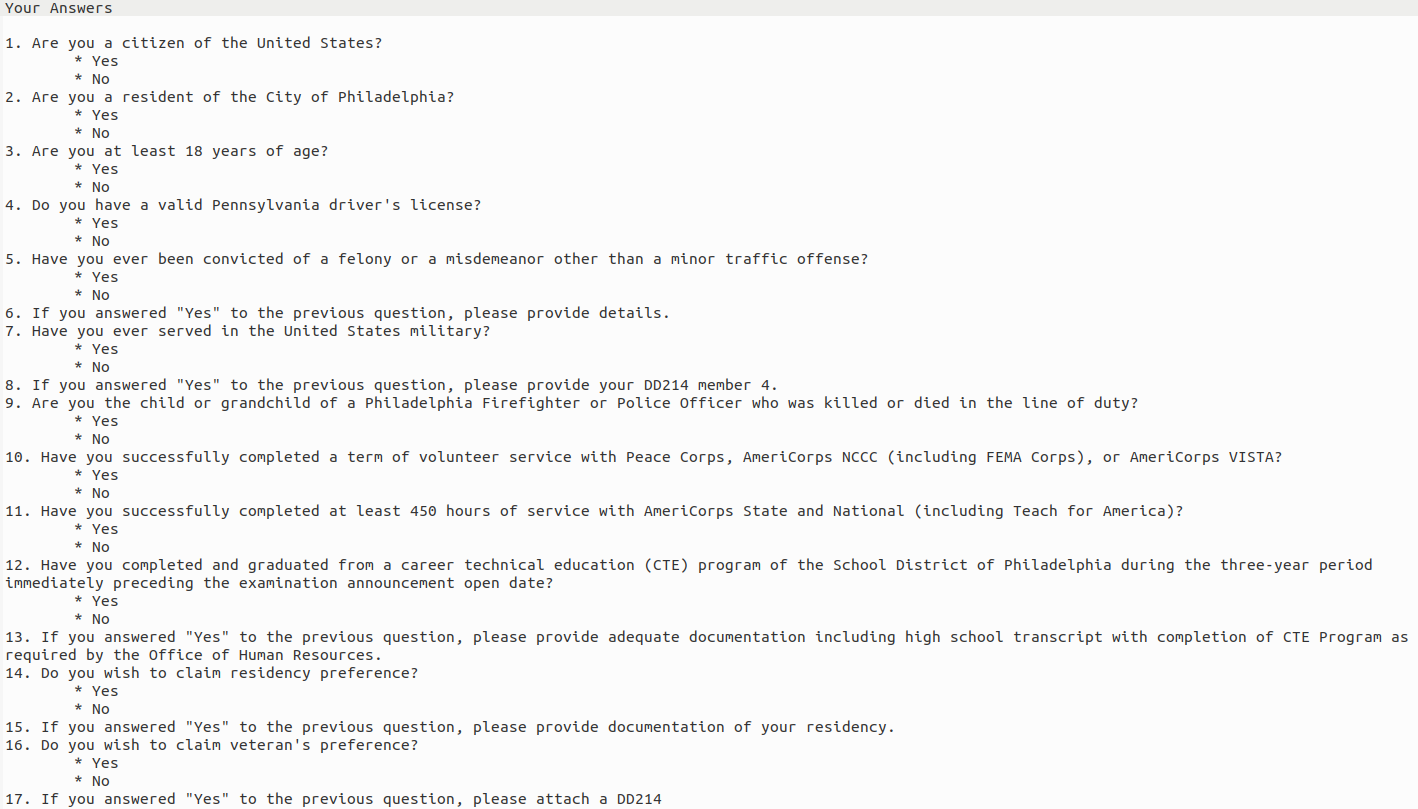}
\caption{Summary for a job posting titled ``Airport Operations Trainee''.}
\label{fig:airport_summary}
\end{figure}

\begin{figure}[!h]
\includegraphics[width=\textwidth]{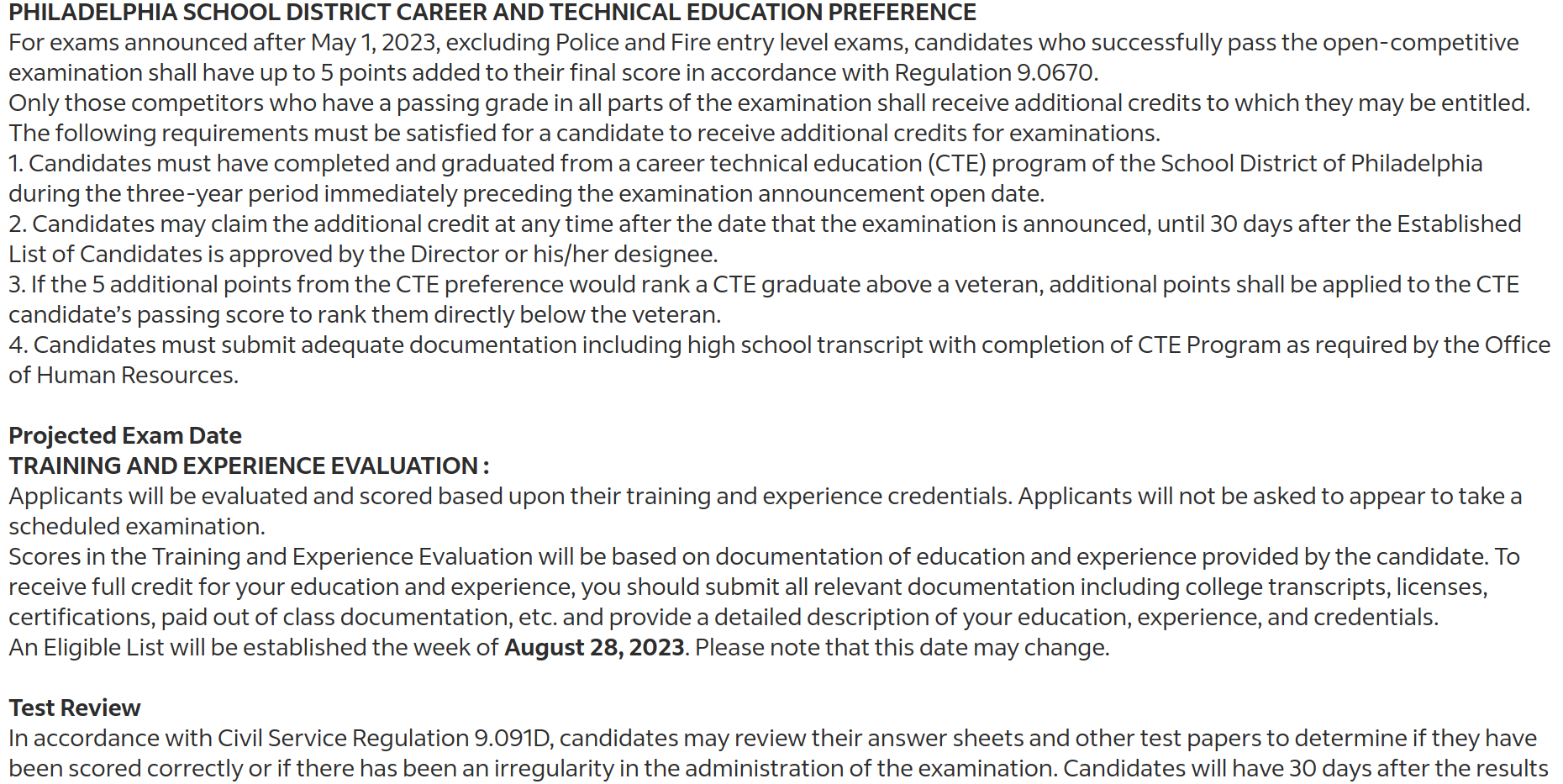}
\caption{Partial content from the ``Airport Operations Trainee'' job posting.}
\label{fig:airport_screenshot}
\end{figure}

\paragraph{Case 1: Member Service Representative}
A job advertisement for a ``Member Service Representative'' presented annotation challenges due to its ambiguous description. Human annotators initially diverged in their classifications, choosing codes 4110.1.1 (Membership administrator) and 4225.1 (Customer service representative). After deliberation, they agreed on 4226.1 (Receptionist). Without summarization, the model's predictions leaned towards 4225, 1412, and 2424. However, the introduction of the summarization step refined these predictions to 4226, 4225, and 1412, placing the consensus label at the forefront. The summary, visualized in Figure \ref{fig:member_summary}, focused on responsibilities typical of receptionists, enhancing the model's accuracy.

\paragraph{Case 2: Procurement Analyst} 
A ``Procurement Analyst'' job posting underwent label revisions from 3323.2 (Purchaser) to 3343.1.5 (Procurement support officer) after discussions among human annotators. Without summarization, the model's top-3 predictions comprised 2424, 3323, and 1219. Introducing summarization realigned these predictions to 3343, 3323, and 4322, with the correct label taking precedence. The summary, depicted in Figure \ref{fig:procurement_summary}, concentrated on core responsibilities, facilitating a precise match.

\paragraph{Case 3: Airport Operations Trainee}
The job posting for an ``Airport Operations Trainee'' posed challenges due to its intricate details. Human annotators debated between 3343.4 (Management assistant) and 3154.1.3 (Airport operations officer) before agreeing on the former. The model, however, failed to predict the correct label, regardless of the summarization step. A closer look at the summary (Figure \ref{fig:airport_summary}) revealed a lack of clarity, possibly influenced by the original posting's detailed exam information (Figure \ref{fig:airport_screenshot}). Refining the summarization prompt or enhancing data preprocessing might address such complexities in future iterations.

\section{Related work}\label{sec:related_work}
The extraction of occupational information from unstructured job postings and resumes has been a longstanding challenge. Various methodologies have been proposed, and in this section, we provide an overview of prominent approaches, contextualizing our own contributions.

\emph{Supervised Approaches}: Supervised learning has conventionally been the dominant paradigm for occupation extraction and standardization. Typically, the extraction task is framed as a multiclass classification problem. Studies such as \cite{boselli2017using,boselli2018wolmis,boselli2018classifying,colombo2019ai} have evaluated the merits of SVMs and convolutional neural networks using English job titles labeled with ISCO codes. In a different vein, \cite{varelas2022employing} employed an ensemble of five machine learning algorithms for classifying Greek job postings into ISCO categories. Other researchers like \cite{bethmann2014automatic} have applied Naive Bayes and Bayesian Multinomial techniques for the prediction of KldB2021 codes using German survey data. Furthermore, investigations such as \cite{russ2016computer} and \cite{mukherjee2021determining} have assessed the efficacy of a variety of supervised algorithms for classifying into American SOC codes. However, supervised methods often grapple with limitations tied to dataset size, language specificity, and taxonomy constraints. Our approach, by contrast, offers a versatile framework, adept at handling multiple languages and taxonomies through a cohesive pipeline.

\emph{Unsupervised Approaches}: Initial endeavors in the realm of occupation and skill extraction leaned on techniques like rule-based labeling \cite{cascot}, keyword-driven searches \cite{laborr}, and topic modeling \cite{gurcan2019big}. An intriguing observation from \cite{wan2023automated} highlighted that CASCOT, a rule-based approach, could sometimes surpass machine learning counterparts in specific evaluations. However, as our results indicate, the performance of such systems still falls short when juxtaposed with our LLM4Jobs framework.

\emph{Transformer-based LLMs for OES}: To the best of our knowledge, LLM4Jobs is the first attempt at harnessing the capabilities of state-of-the-art GPT-style decoder-only models for occupation coding. While there exist some parallels in the HR AI domain, such as skill extraction, the focus has primarily been on BERT-like (encoder-only) architectures \cite{devlin2018bert,liu2019roberta}. For instance, \cite{vermeer2020using} limited their scope to fine-tuning classification layers for skill categorization. Meanwhile, works like \cite{chernova2020occupational,zhang2022kompetencer} leveraged LLM embeddings as alternatives to traditional word2vec embeddings, albeit with the LLM playing a somewhat peripheral role in their Named Entity Recognition (NER) frameworks. Predominantly, prior endeavors have concentrated on tapping into LLMs for Natural Language Understanding (NLU). LLM4Jobs covers new grounds by exploring the generative prowess (NLG) of LLMs specifically for OES.

\section{Conclusion}\label{sec:conclusion}
We introduced LLM4Jobs, an adaptable, unsupervised large language model based approach for occupation extraction and standardization. Through extensive experimentation, LLM4Jobs consistently outperformed state of the art unsupervised methods, showcasing its robustness across varied datasets and granularity levels.
LLM4Jobs is the first framework for occupation coding task that leverages the natural language generation capacities of LLMs.  Given the lack of benchmarks, we introduced synthetic and real-world datasets, providing a foundation for future research. LLM4Jobs's reliance on open-sourced LLMs ensures cost-effectiveness without sacrificing performance.
There are some limitations. Our real world dataset has a small size due to limitation of resources, and the synthetic datasets might not always mimic real-world nuances, which could reduce the confidence in the evaluation results. Some popular options for optimizing the performance of large language models, such as fine-tuning and prompt engineering, have not properly been examined yet.
In the future, we plan to expand and evaluate LLM4Jobs's applicability to diverse taxonomies. 
Assessing the potential of fine-tuning existing models with occupation-centric data is another promising direction.
To conclude, LLM4Jobs stands as a promising step in unsupervised automatic occupation coding, yet acknowledging its limitations paves the way for future enhancements.

\section{Acknowledgements} The research leading to these results has received funding from the European Research Council under the European Union's Seventh Framework Programme (FP7/2007-2013) (ERC Grant Agreement no. 615517), and under the European Union’s Horizon 2020 research and innovation programme (ERC Grant Agreement no. 963924), from the Flemish Government under the ``Onderzoeksprogramma Artificiële Intelligentie (AI) Vlaanderen'' programme, and from the FWO (project no. G0F9816N, 3G042220).

During the preparation of this work the authors used ChatGPT in order to improve language and readability only. After using this tool, the authors reviewed and edited the content as needed and take full responsibility for the content of the publication.
\bibliographystyle{splncs04}
\bibliography{ref}

\appendix

\section{Prompts for GPT4}\label{app:gpt4_prompts}
The prompt for generating the easy synthetic dataset is as follows.
\begin{verbatim}
You are an expert in HR and ISCO/ESCO classification.
Your task is to write a sample job posting that fits
the given 1 or more ESCO occupation codes.
The job posting must NOT mention the code and 
be less than 250 words. Begin ESCO codes:
\end{verbatim}

For the hard dataset:
\begin{verbatim}
You are an expert in HR and ISCO/ESCO classification.
Your task is to write a sample job posting that fits
the given 1 or more ESCO occupation codes.
The job title must be different from the occupation.
The job posting must NOT mention the code and 
be less than 250 words. Begin ESCO codes:
\end{verbatim}

The prompt for zero-shot prediction is as follows.
\begin{verbatim}
As an experienced HR expert, you have been tasked with
analyzing job descriptions and assigning them 
appropriate ISCO/ESCO occupation codes. 
Based on your expertise and understanding of the ISCO/ESCO
classification system, please assign the most appropriate 
**10** occupation codes to these job descriptions.
ONLY answer with the codes and the corresponding titles 
of the occupations, e.g. 1120.3.1\tairport chief executive, 
in tsv format of 10 lines with column header: 
code\ttitle and with separator: \t. 
Do NOT elaborate. Do NOT explain. Do NOT return more than 11 lines.
\end{verbatim}

\section{Configurations for LLM4Jobs summarization}\label{app:generation_hyperparam}
For summarization, we set temperature to be 0., and use the following prompt:

\begin{verbatim}
Human: Given the following job posting, please summarize 
the key duties and functionalities. 
Your summary should be within 10 bullet points of 
duties/functionalites/responsibilities. 
Here is an example summary with the required format. 
You need to follow the template below.

Job Title: ATM repair technician
Key responsibilities:
1. install, diagnose, maintain and repair automatic teller machines. 
2. travel to clients' location to provide their services. 
3. use hand tools and software to fix malfunctioning money distributors.
4. night shifts.

Now begins the job posting you need to summarize.
\end{verbatim}

\section{Real job posting annotation}\label{app:real_data}
\subsection{Data collection}
We began with using the population of the cities as sampling weights to sample US cities. Then for each of the sampled cities (repetition existing), we randomly sampled one job posting from the first 3 pages of the results from searching jobs located at that city, until 100 samples are collected.

\subsection{Annotation}
There are three researchers serving as annotators. 
Sixty postings were divided into three sets of 20, each annotated by a pair of researchers. The rest forty are annotated by two researchers. In instances of disagreement, a third researcher, unfamiliar with the sample, arbitrated. 
The initial annotation round surfaced disagreements on 30 labels up to level-4 and 48 labels up to the most detailed level, underscoring the task's inherent complexity. Post-discussion, while consensus was reached on all labels, a few remained contentious, as exemplified below.

\subsection{Tricky cases}\label{app:tricky_jds}
Below are the screenshots of the job postings we annotators find hard to classify.

\begin{figure}
\includegraphics[width=\textwidth]{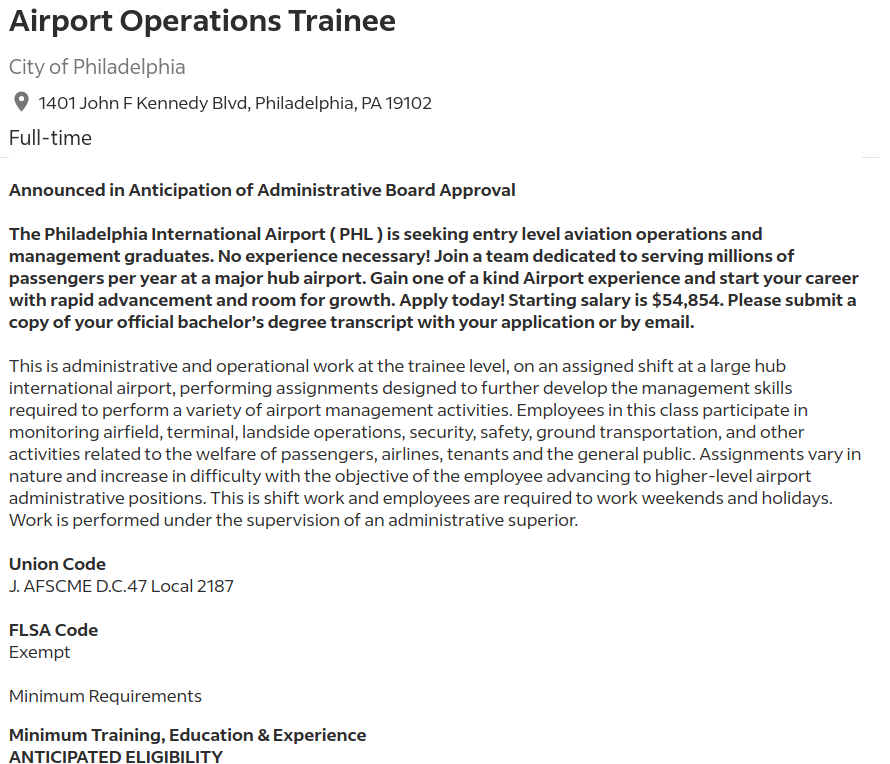}
\label{fig:airport_trainee}
\caption{Airport Operations Trainee job posting (partial) from \url{https://www.indeed.com/viewjob?jk=ab5b16b11220e958&from=serp&vjs=3}. The agreed code is 3343.4 management assistant.}
\end{figure}

\begin{figure}
\includegraphics[width=\textwidth]{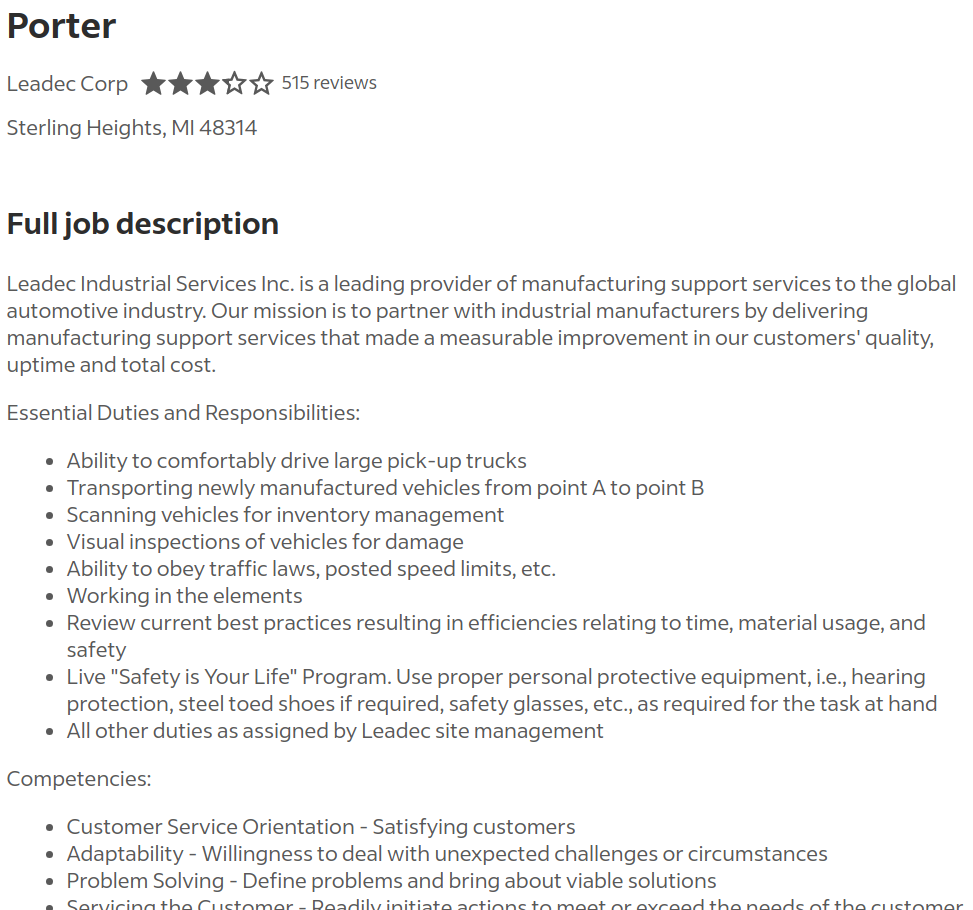}
\label{fig:porter}
\caption{Porter job posting (partial) from \url{https://www.indeed.com/viewjob?jk=f7f8960dd560cb46&from=serp&vjs=3}. The agreed code is 9333.4 mover.}
\end{figure}

\begin{figure}
\includegraphics[width=\textwidth]{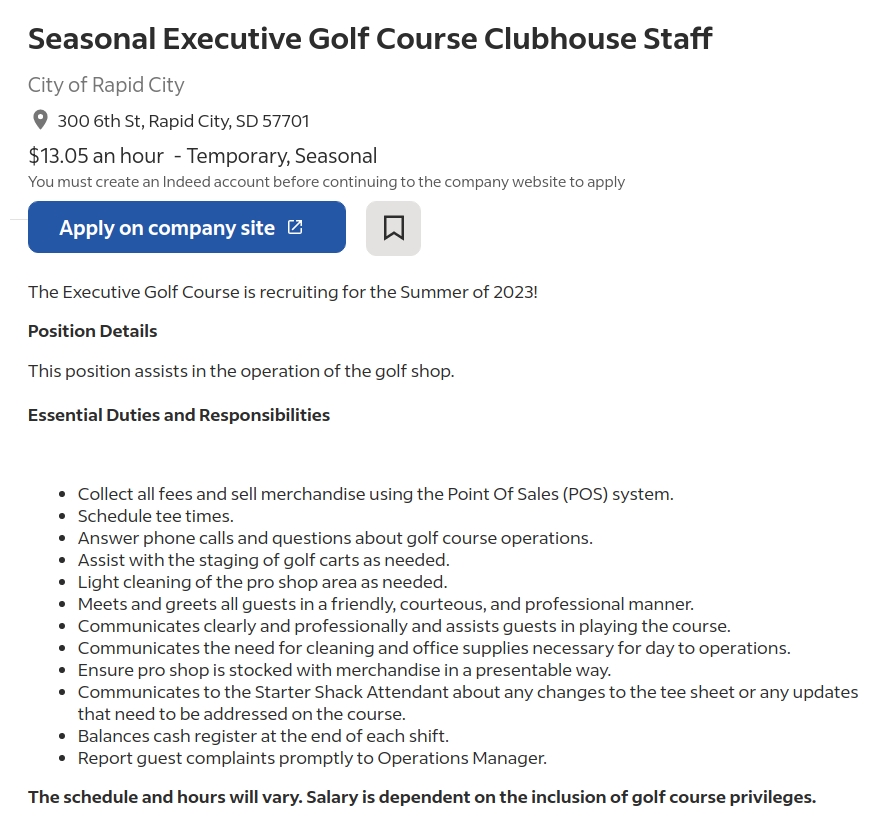}
\label{fig:glofclub_staff}
\caption{Seasonal Executive Golf Course Clubhouse Staff job posting (partial) from
\url{https://www.indeed.com/viewjob?jk=d153d056c2ed9839&from=serp&vjs=3}. The agreed code is 5223.6 shop assistant.}
\end{figure}

\begin{figure}
\includegraphics[width=\textwidth]{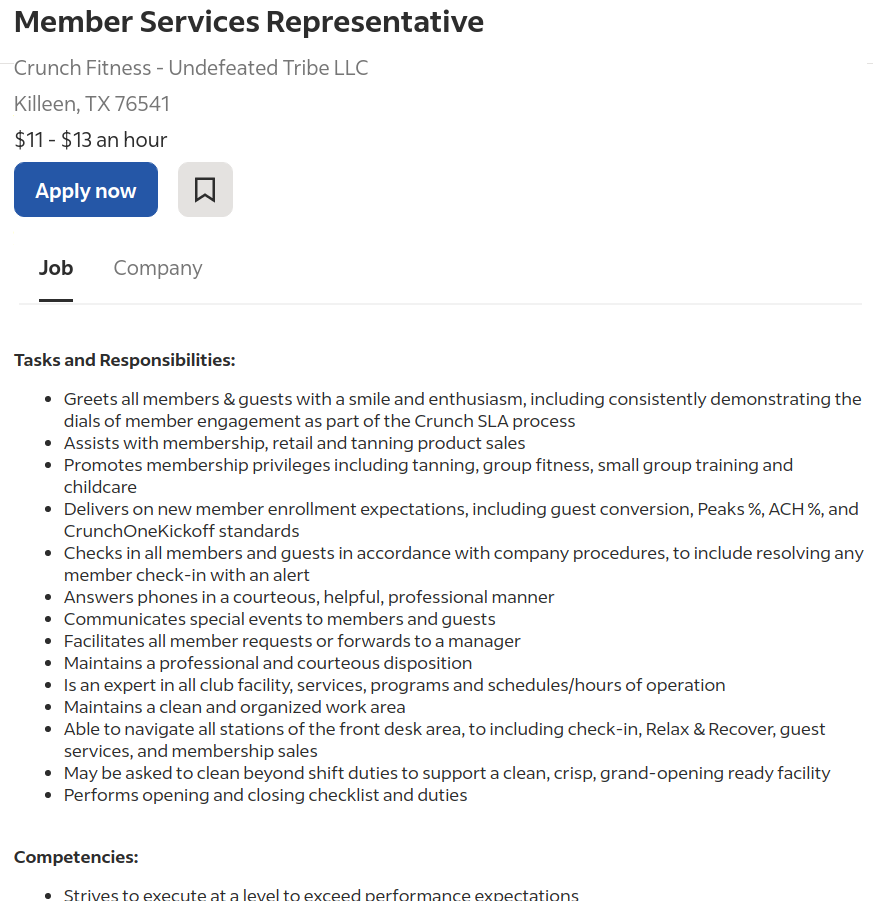}
\label{fig:member_repr}
\caption{Member Services Representative job posting (partial) from \url{https://www.indeed.com/viewjob?jk=9b09353a4f725a7a&from=serp&vjs=3}. The agreed code is 4226.1 receptionist.}
\end{figure}

\begin{figure}
\includegraphics[width=\textwidth]{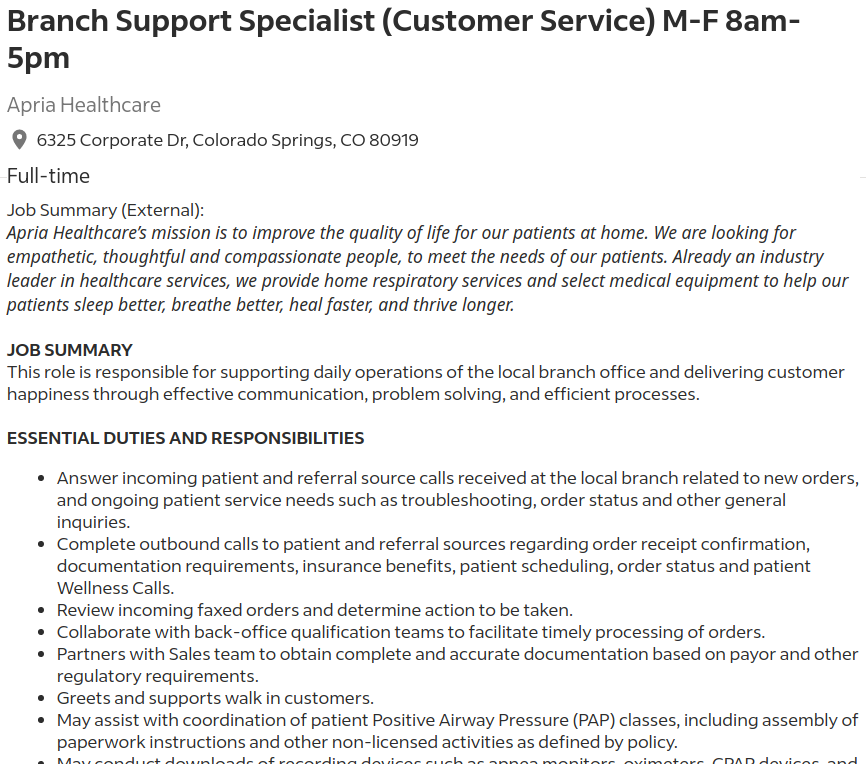}
\label{fig:branch_support}
\caption{Branch Support Specialist job posting (partial) from \url{https://www.indeed.com/viewjob?jk=df4b2f8a073bdc3f&from=serp&vjs=3}. The agreed code is 5223.4 sales assistant.}
\end{figure}

\end{document}